\documentclass[final,5p,times,twocolumn]{elsarticle}

\usepackage[utf8]{inputenc}
\usepackage{newunicodechar}
\newunicodechar{ℓ}{l}

\usepackage{lineno}
\usepackage{graphicx}
\usepackage{amssymb}
\usepackage{pdflscape}
\usepackage[english]{babel}
\usepackage[dvipsnames,svgnames,x11names]{xcolor}
\usepackage{booktabs,tabularx}
\usepackage{multirow}
\usepackage{subfig} 
\usepackage{hyperref}
\usepackage{amsmath}
\usepackage{commath}
\usepackage[english]{babel}
\usepackage[ruled]{algorithm2e}
\SetEndCharOfAlgoLine{}
\usepackage[leftmargin=6em,rightmargin=12em,indentfirst=false]{quoting}
\usepackage{siunitx}

\usepackage[final]{changes}
\usepackage{float}

\usepackage{lineno}
\usepackage{tikz}
\usepackage[export]{adjustbox}

\usepackage{graphicx}
\usepackage[utf8]{inputenc}
\usepackage[export]{adjustbox}
\usepackage{wrapfig}
\usepackage{dcolumn}

\makeatletter
\newcolumntype{T}[3]{>{\textfont0=\the@{#1}{#2}{#3}}c<{\DC@end}}
\makeatother

\usepackage{pgfplots}
\pgfplotsset{width=10cm,compat=1.9}

\usepackage{array}

\newcolumntype{L}[1]{>{\raggedright\let\newline\\\arraybackslash\hspace{0pt}}m{#1}}
\newcolumntype{C}[1]{>{\centering\let\newline\\\arraybackslash\hspace{0pt}}m{#1}}
\newcolumntype{R}[1]{>{\raggedleft\let\newline\\\arraybackslash\hspace{0pt}}m{#1}}

\usepackage{subfiles}

\usepackage{todonotes}

\journal{Applied Energy}
\begin{document}
	
\begin{frontmatter}

\title{Using Google Trends as a proxy for occupant behavior to predict building energy consumption}

\author{Chun Fu and Clayton Miller\,$^{*}$}

\address{Department of the Built Environment, National University of Singapore (NUS), Singapore}

\address{$^*$Corresponding Author: clayton@nus.edus.sg, +65 81602452}

\begin{abstract}
In recent years, the availability of larger amounts of energy data and advanced machine learning algorithms has created a surge in building energy prediction research. However, one of the variables in energy prediction models, occupant behavior, is crucial for prediction performance but hard-to-measure or time-consuming to collect from each building. This study proposes an approach that utilizes the search volume of topics (e.g., \emph{education} or \emph{Microsoft Excel}) on the Google Trends platform as a proxy of occupant behavior and use of buildings. Linear correlations were first examined to explore the relationship between energy meter data and Google Trends search terms to infer building occupancy. Prediction errors before and after the inclusion of the trends of these terms were compared and analyzed based on the ASHRAE Great Energy Predictor III (GEPIII) competition dataset. The results show that highly correlated Google Trends data can effectively reduce the overall RMSLE error for a subset of the buildings to the level of the GEPIII competition's top five winning teams' performance. In particular, the RMSLE error reduction during public holidays and days with site-specific schedules are respectively reduced by 20-30\% and 2-5\%. These results show the potential of using Google Trends to improve energy prediction for a portion of the building stock by automatically identifying site-specific and holiday schedules.
\end{abstract}

\end{frontmatter}


\section{Introduction}

Building energy prediction has been an important research topic in recent decades due to its extensive use in thermal load prediction~\cite{Wang2020-ky}, systems optimization~\cite{Fan2020-tu,Li2021-wb}, measurement and verification~\cite{Granderson2015-ms,Granderson2016-wq, Granderson2017-lm}, calibration~\cite{Chong2017-mk,Chong2019-xh}, and retrofit analysis~\cite{Deb2021-py}. A recent text-mining study of over 30,000 publications related to building energy efficiency and data science showed the growth of prediction techniques over the last decade~\cite{Abdelrahman2021-bk}. There are many types of building energy prediction methods, including traditional statistical time-series models~\cite{Kawashima1995-aw, Ruch1993-bh,Gunay2017-ke}, building simulation models (e.g., EnergyPlus)~\cite{Neto2008-bo}, and some popular machine learning models such as neural networks~\cite{Moon2019-oy, Ahmad2017-jz,Li2021-wb}, deep learning~\cite{Wang2020-ky,Brandi2020-ea,Nichiforov2018-ne,Fan2017-ac}, or tree-based models~\cite{Touzani2018-lz}. Among them, machine learning models are currently receiving the most attention because of their prediction accuracy and multi-factor flexibility~\cite{Amasyali2018-wj}. For example, in the overview paper of the Great Energy Predictor III (GEPIII) machine learning competition\footnote{\url{https://www.kaggle.com/c/ashrae-energy-prediction}} hosted by the ASHRAE organization on the Kaggle platform~\cite{Miller2020-fo}, it was highlighted that most of the winning teams used gradient boosting tree-based models with model-ensembling techniques to predict building energy with meta, weather, and temporal data. This competition was the latest generation of the series of ASHRAE-led competitions~\cite{Haberl1998-du,Kreider1994-dn,Katipamula1996-et,Haberl1998-du,Ohlsson1994-vl} towards creating a machine learning body of knowledge, open data set~\cite{Miller2020-yc}, and means of benchmarking energy prediction methods~\cite{Miller2019-sg}. This momentum demonstrated considerable research potential in the field of building energy forecasting.

\begin{figure*}
\begin{center}
\includegraphics[width=0.9\textwidth, trim= 0cm 0cm 0cm 0cm,clip]{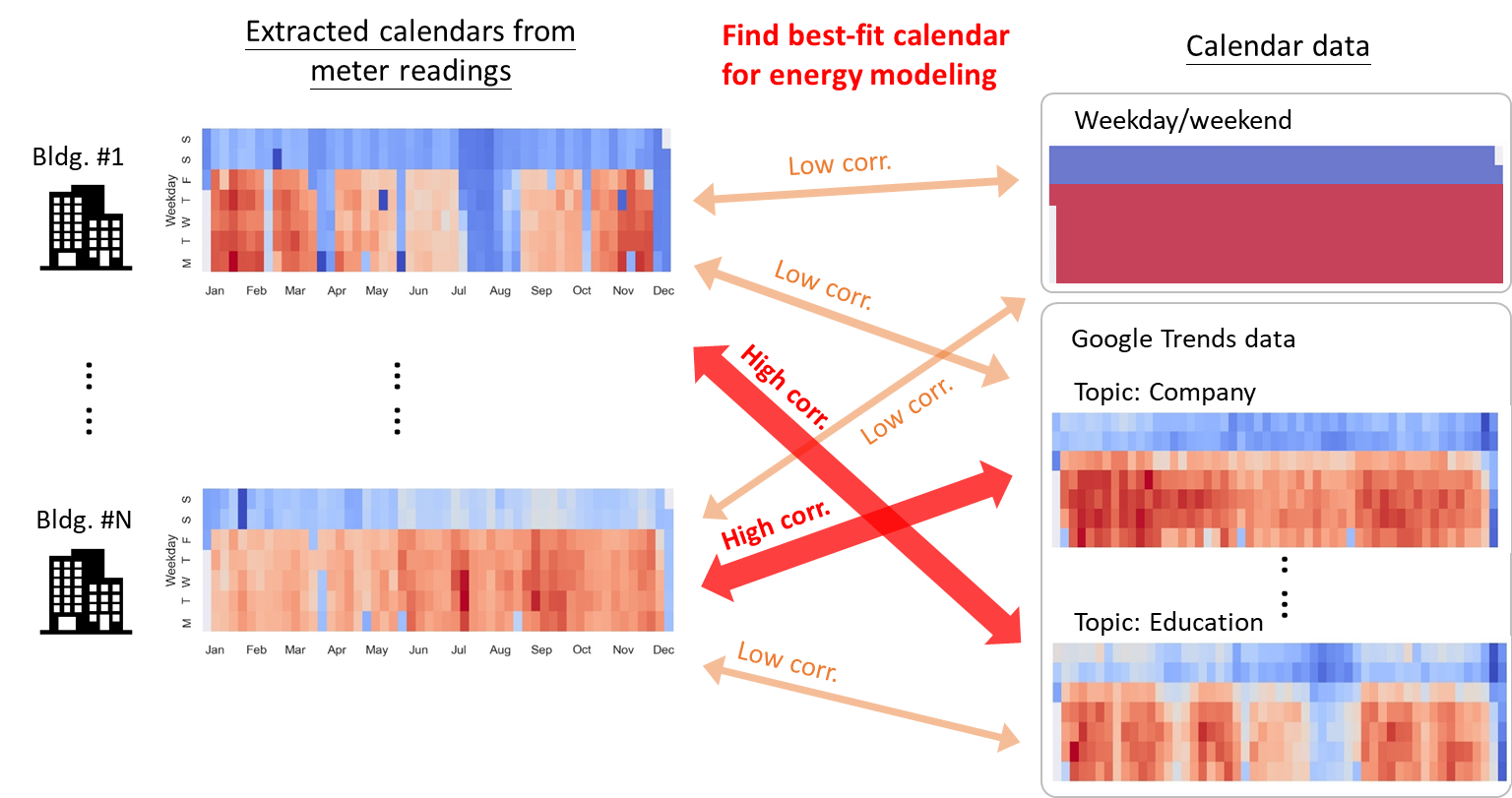}
\caption{Correlation between data sources like weekday, weekend, and holidays schedules and building energy consumption is well understood. The use of Google Trends as a means of detecting occupant behaviour could automate the process of characterizing the localized effects of unique site-specific schedules.}
\label{fig:overview}
\end{center}
\end{figure*}

\subsection{Occupant behavior and patterns data could improve prediction}
Although there has been considerable research in building energy prediction,  a common challenge in building energy modeling is the lack of occupant behavior in the dataset, which may negatively affect prediction performance. An error analysis of the GEPIII competition showed that the average prediction of the top 50 teams in the competition had a higher than acceptable range of error in over 20\% of the test data set~\cite{Miller2021-bf}. The error regions were especially high for hot water meters (60\% of test data) and primary use types like science/technology (45.2\% of test data). One of the key reasons for these errors is challenges in creating occupancy schedules during periods of localized holidays, break periods, time off, or other effects~\cite{Touzani2019-sz}. University campuses, for example, have periods such as \emph{spring break} in which students are not in class, which results in energy usage being different from regular days - usually lowered. Further, the site-specific calendars of each building and sites can be different, making it difficult to scale the use of such data. These schedules from different locations require additional manual work to collect and organize. In terms of modeling, one strategy is to separately train models by day types, such as regular-day and holiday energy models, but there might be a higher risk of error due to insufficient training data. Some studies combine fuzzy models and similar day methods for training~\cite{Wi2012-rv, Ebrahimi2013-pk, Ma2017-vk}. Some studies use holidays as the input by binary encoding (i.e., 0 for a non-holiday, and 1 for a holiday), but efforts in manually collecting calendar data and confirming the building type are inevitable~\cite{Dahl2018-ts, Zhang2019-dr}. Due to the tedious efforts to collect calendar data, most past research focused on buildings in a single region, and no research has yet to propose a general method that can be widely applied across building types or countries.

\subsection{Possible use of Google Trends as a proxy for building occupancy}
Due to the popularity of mobile devices and networks, the access of such mobility data combined with data science is a promising research direction. Google Trends, a popular search volume query platform based on the most popular search engine worldwide, with a large enough volume of search,  provides data for understanding what search terms people use to find information on the Internet\footnote{\url{https://trends.google.com/trends/}}. Examples of work using Google Trends as a means of improving the prediction of human-related behavior include papers in health care~\cite{Zhang2018-av, Nuti2014-bu,  Mavragani2020-yp}, financial markets~\cite{Preis2013-du, Woo2018-ej, Carriere-Swallow2013-fu, Vosen2011-oo}, and tourism~\cite{Clark2019-bg, Onder2017-jo, Dergiades2018-ml}. In the building performance field, Google's Popular Time was used to create weekly data-driven schedules for each building type and compared with standard schedules from ASHRAE~\cite{Happle2020-el}. In other studies, crowd positioning data were used to extract occupancy patterns in buildings~\cite{Barbour2019-sc, Kang_undated-hd}. However, in the research of building energy prediction, the state-of-the-art does not leverage any online data beyond manually collected and organized data on weekdays/weekends and national holidays. Therefore, this work uses the search volume provided by the Google Trends data in an innovative way to improve the energy model.

\subsection{Research objectives and novelty}
Figure \ref{fig:overview} shows the goal of this paper in its focus on using data collected from Google Trends to find correlating signals that can help predict energy consumption for various buildings. Google Trends provides a search volume of specific topics with selected time periods and regions, representing the trend of people's behavior and providing calendars for the building energy model. The purpose of this research is to provide a framework, starting from selecting and evaluating topics, verifying whether topics can improve the accuracy of the energy model across buildings and countries, and finally suggesting how to use Google Trends data to improve the prediction model. This study raises the following three questions:

\begin{enumerate}
\item How to evaluate whether the search volume of topics matches the energy use behaviors of the building?
\item How much accuracy of the energy model can be improved/reduced by provided calendars from Google Trends?
\item What are the topics that can improve the performance of model prediction?
\item Could this method be scaled across multiple sites from a diversity of locations?
\end{enumerate}

The rest of the paper is organized as follows. Section \ref{sec:methods} outlines the process of collecting and cleaning the data, performing a correlation analysis to filter the best topics for prediction, and testing the ability of Google Trends to influence prediction accuracy. Section \ref{sec:results} shows the effect of implementing the proposed method on a large set of data from energy meters in the GEPIII. Finally, Sections \ref{sec:discussion} and \ref{sec:conclusion} interpret the results in the context of improving the field of energy prediction in buildings and indicates the limitations of this paper.

\begin{figure}
\begin{center}
\includegraphics[width=0.45\textwidth, trim= 0cm 0cm 0cm 0cm,clip]{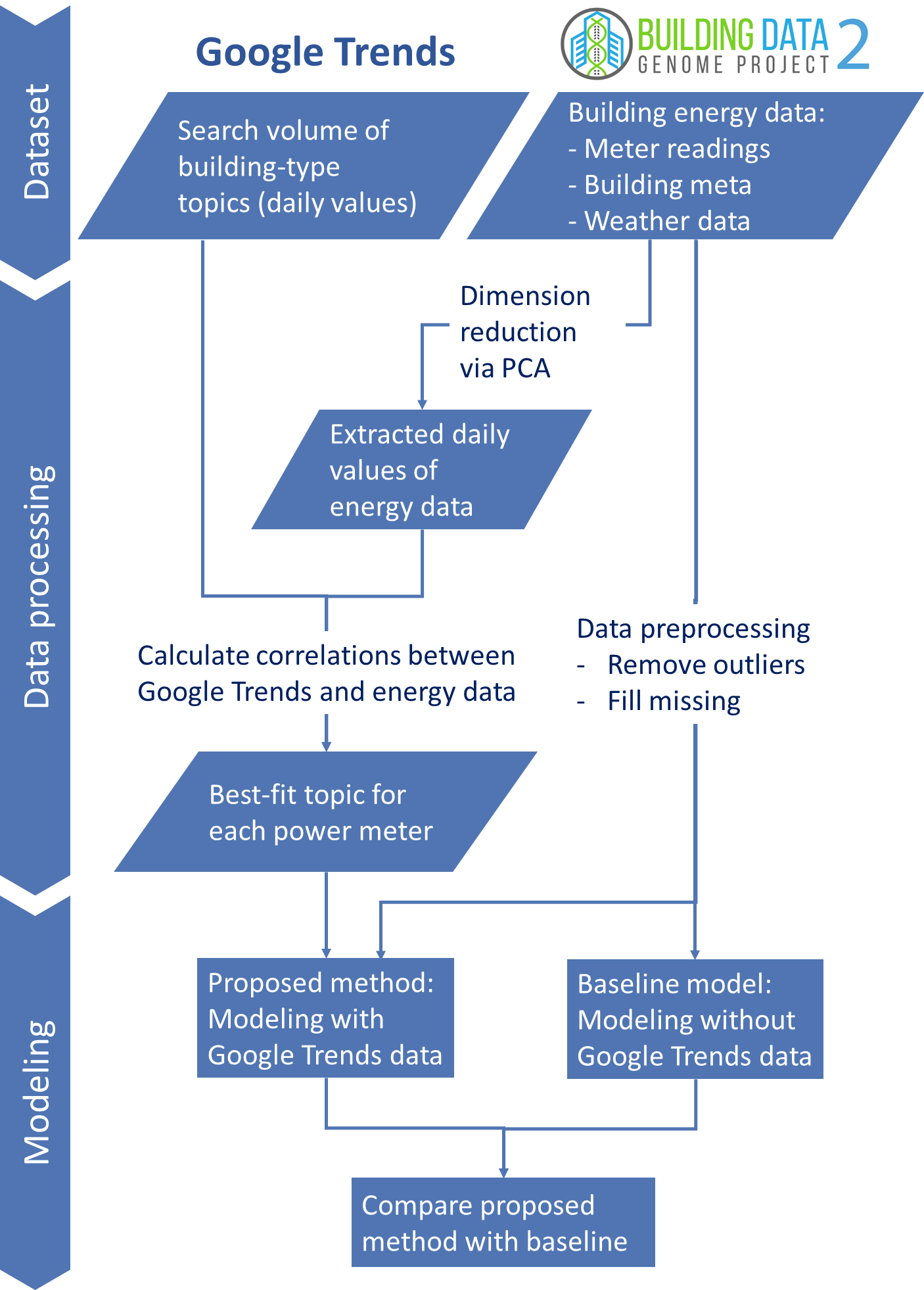}
\caption{Overview of the methodology to process data from the BDG2 and Google Trends data sources to test prediction improvement over the GEPIII competition-inspired baseline.}
\label{fig:workflow}
\end{center}
\end{figure}

\section{Methodology}
\label{sec:methods}

For this study, the time-series hourly data from 2,380 meters in the Building Data Genome 2 (BDG2) project dataset were used for modeling. These data were also used in the GEPIII Kaggle competition~\cite{Miller2020-fo}. Figure \ref{fig:workflow} illustrates the flow of the framework, starting from preprocessing energy and Google Trends data, selecting and evaluating topics, and finally, comparing the proposed method with the baseline model. 

\subsection{Datasets}
Two datasets were used in this study: (1) building energy data and (2) Google Trends data to evaluate the impact of search volumes on the performance of building energy models. Both datasets contain time-series data from 2016 to 2017. For this study, the data from 2016 was used as training, and the data from 2017 was retained as validation.

\subsubsection{Building energy meter data}
The Building Data Genome 2.0 (BDG2) is an open dataset containing hourly meter readings and metadata of 3,053 power meters over two years. Each of the buildings has metadata such as floor area, weather, and primary use type. This dataset can serve as a benchmark for comparing different machine learning algorithms and data science techniques. This study compared the prediction result before and after adding Google Trends based on the baseline model in GEPIII, so only 2380 meters out of 3053 meters were used in this study to be consistent with the competition settings. Table \ref{tab:metadata} outlines the metadata variables available in the data set.

\begin{table*}
\centering
\resizebox{0.9\textwidth}{!}{%
\begin{tabular}{llll}
\hline
\textbf{Category} & \textbf{Variable} & \textbf{Unit} & \textbf{Content} \\ \hline
Power   meters & Meter type & - & Type of meter: electricity, chilled   water, steam, or hotwater \\
 & Meter readings & kWh & Energy consumption in \\
 &  &  &  \\
Building meta & Primary use & - & Primary category of activities for the building \\
 & Year built & - & Year building was opened \\
 & Floors & - & Number of floors of the building \\
 & Floor area & Square foot & Gross floor area of the building in square feet \\
 &  &  &  \\
Weather data & Temperature & Degree Celsius & Outdoor temperature \\
 & Cloud cover & Oktas & Portion of the sky covered \\
 & Dew point & Degree Celsius & Outdoor dew temperature \\
 & Precipitation & Millimeter & Precipitation depth \\
 & Pressure & Millibar & Sea level pressure  in \\
 & Wind speed & m/s & Wind speed in \\
 & Wind bearing & Degree (0-360) & Wind direction \\ \hline
\end{tabular}
}
\caption{Building metadata available in the Building Data Genome 2 (BDG2) project.}
\label{tab:metadata}
\end{table*}

\subsubsection{Google Trends data}

\emph{Google Trends} is a service that analyzes the popularity of search queries in various regions and languages in Google search. It provides users with normalized time series of search volumes (between 0 and 100) for any keywords of interest. On the Google Trends platform, there are two sources of search volumes: \emph{topics} and \emph{terms}. In this research, only search volumes of \emph{topics} were collected and applied to the energy model. The reason for using \emph{topics} instead of \emph{terms} is that definition of  \emph{topics} in Google Trends is a set of terms that share the same concept across languages, while \emph{terms} only show matches in user's query in the given language. For instance, the official example gives that the research volume of the \emph{topics} \emph{London} includes results such as \emph{British Capital} and \emph{Londres} (Spanish for \emph{London}). But the \emph{terms} \emph{London} only contains the words \emph{London} or \emph{London Bridge}, and does not contain words in other languages or related concepts. Therefore, using \emph{topics} instead of \emph{terms} better captures a set of keywords and be well applied in different countries. 

\begin{figure}
\begin{center}
\includegraphics[width=0.45\textwidth, trim= 0cm 0cm 0cm 0cm,clip]{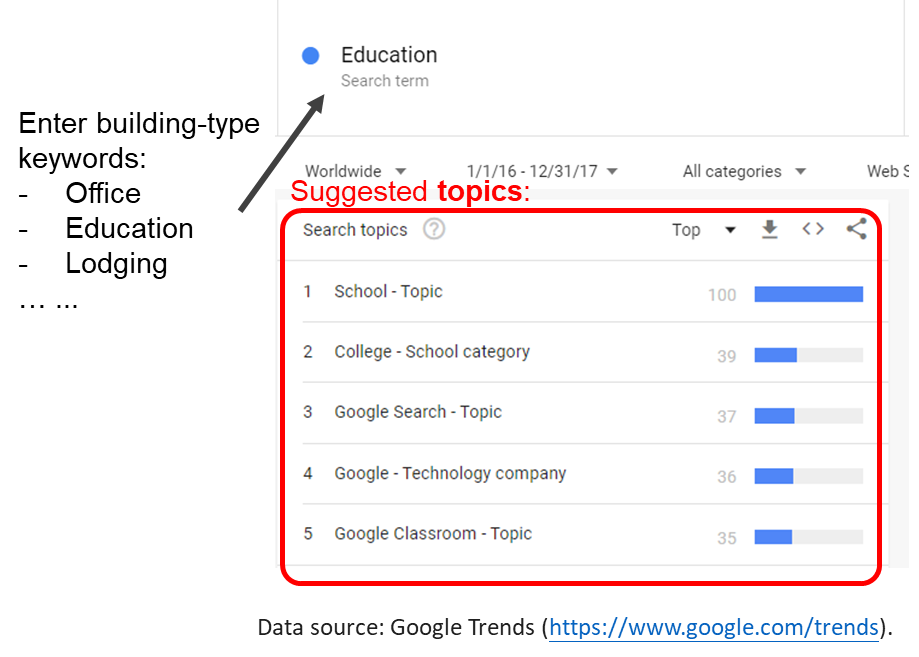}
\caption{Example of finding and extracting the suggested topics from the Google Trends interface.}
\label{fig:gettingtopics}
\end{center}
\end{figure}

For the selection of topics, in previous studies, search terms and topics for modeling were manually collected based on the author's knowledge and judgment~\cite{Preis2013-du, Mavragani2020-yp, Zhang2018-av, Woo2018-ej, Onder2017-jo}. However, out of simplicity and intuition, this research directly uses the \emph{primary use} column from dataset as initial search keywords (e.g., \emph{Education}, \emph{Office}, \emph{Retail}, etc.) to collect appropriate topics from suggested ones (see Figure \ref{fig:gettingtopics}). A similar selection process was also seen in a study predicting trading behaviour~\cite{Preis2013-du}, which used terms suggested by the \emph{Google Sets} service by giving keywords related to the stock market. For example, if entering \emph{education} as search term on the platform, the suggested topics include \emph{Education}, \emph{Higher education}, \emph{Sex Education} (A comedy-drama series), etc. Out of these suggested topics, only reasonable ones were selected, and irrelevant topics like \emph{Sex Education} would be therefore removed. Furthermore, additional keywords of productivity tools were also considered: \emph{Microsoft Office}, \emph{Google Docs}, and \emph{Mails}. Because these tools are often used in the workplace, derived topics are expected to be related to human behavior in buildings (e.g., looking up "how do I do vlookup on Microsoft excel" on Google Search). Table \ref{tab:googletrendstopics} shows a table of extracted topics and the categories to which they belong.

\begin{table}
\centering
\resizebox{0.45\textwidth}{!}{%
\begin{tabular}{ll}
\hline
\textbf{Category} & \textbf{Topics of   Google Trends} \\ \hline
\begin{tabular}[c]{@{}l@{}}Building types\\ (\emph{primary use} column\\ from metadata)\end{tabular} & \begin{tabular}[c]{@{}l@{}}Education,   School, Secondary school, \\ High school, School district, \\ Primary school,   Middle school, \end{tabular} \\
 & Lodging,   Boarding house, \\
 & \begin{tabular}[c]{@{}l@{}}Enterprise,   Management, Career, \\ Company, Employment,\end{tabular} \\
 & Retail,   Shopping mall, \\
 & Parking   lot, Parking, \\
 & Public   service, \\
 & Warehouse, \\
 & Place   of worship, \\
 & Health   Care, \\
 & Utility, \\
 & Technology,   Organization, \\
 & Manufacturing, \\
 & Residential   building, \\
 & Storage, \\
 & Science \\
 &  \\
Office productivity tools & Microsoft   Office, Microsoft Outlook, \\
 & Microsoft Excel, Office365, \\
 & \begin{tabular}[c]{@{}l@{}}Google   Docs, Google, Gmail, Google \\ Classroom, Google Drive,\end{tabular} \\
 & Email \\ \hline
\end{tabular}
}
\caption{Overview of Google Trends topics identified for use in this analysis.}
\label{tab:googletrendstopics}
\end{table}

\begin{figure}
\begin{center}
\includegraphics[width=0.45\textwidth, trim= 0cm 0cm 0cm 0cm,clip]{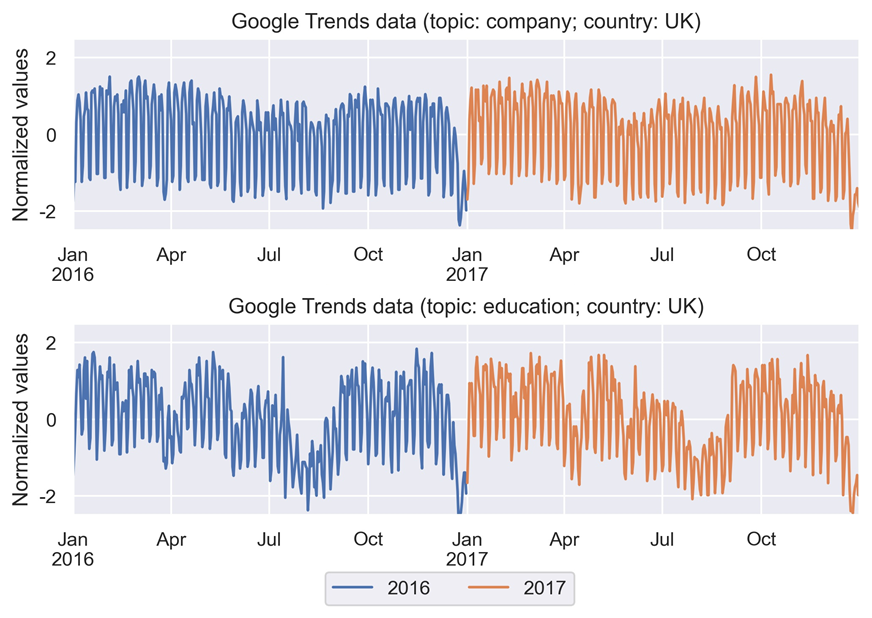}
\caption{Examples of processed data extracted from Google Trends for two topics.}
\label{fig:googletrendsexample}
\end{center}
\end{figure}

The daily search volume of these topics in each country was downloaded in batches according to the selected topics. These daily search volume data included 39 topics from 2016 to 2017 in 4 countries. This study used an open-source Python package called \emph{pytrends} to extract the data set for use in this analysis. To normalize the search volume of different topics in different years in the model, the search volume of all topics was standardized by year (i.e., subtracted by yearly average and divided by yearly standard deviation). Figure \ref{fig:googletrendsexample} visualizes two examples of processed time series for two topics to show the typical patterns and seasonality of such data that is similar to energy consumption measurement.


\subsection{Data preprocessing}
Before developing the energy prediction model, the preprocessing of the data was done to prepare for the analysis. First, due to the daily granularity of Google Trends data, the energy data was converted from hourly to daily values. Next, the best-fit topics for each meter's energy behavior were found via calculating the correlation between meter readings and Google Trends, and then the signals of these topics were used as input features for the prediction model in the next phase.

\subsubsection{Calendar extraction from meter readings}
\begin{figure}
\begin{center}
\includegraphics[width=0.45\textwidth, trim= 0cm 0cm 0cm 0cm,clip]{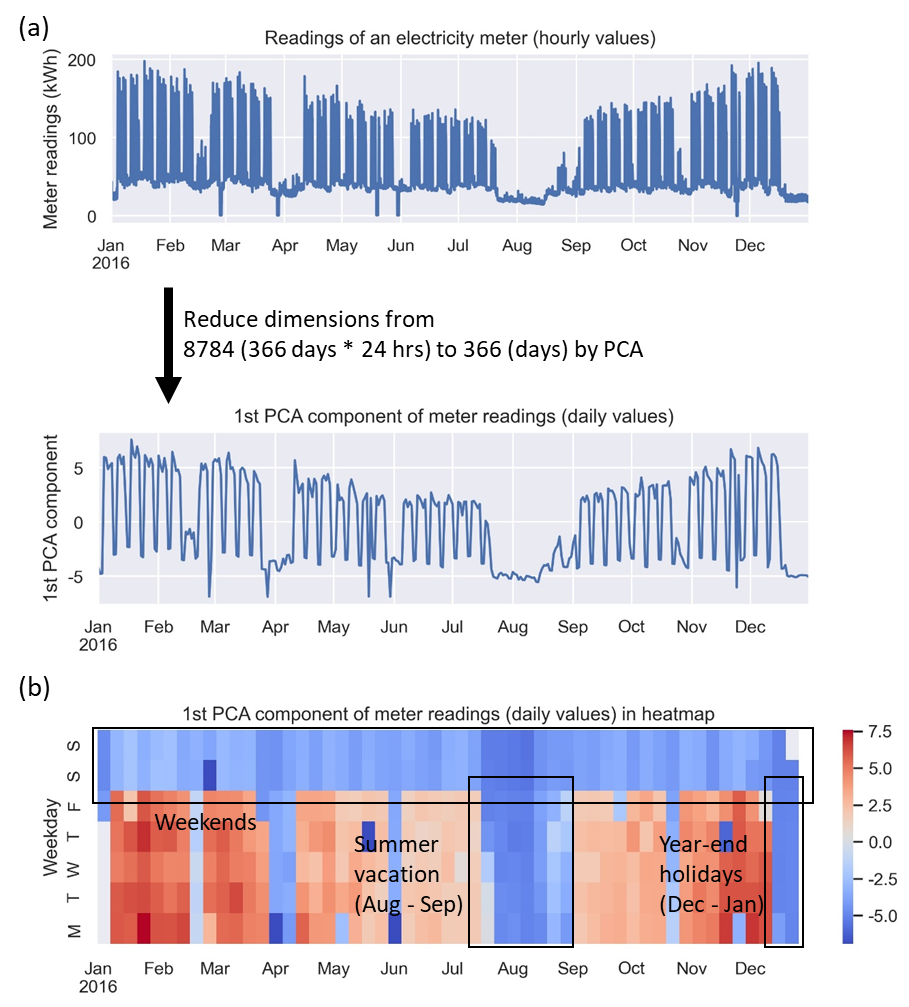}
\caption{An example of the preprocessing of the meter data using the first PCA component.}
\label{fig:preprocessing_meters}
\end{center}
\end{figure}

To be consistent with the daily search volume provided by Google Trends, the hourly meter data in this study was resampled to a daily resolution, which represents the \emph{calendar data} of the meter. Among various techniques of dimension reduction, PCA (Principal component analysis) is a computationally effective way to compress and represent data in lower dimensions via finding principals with the highest variance~\cite{wold1987principal}. This study used PCA to reduce the dimensionality of the annual data from 8784 dimensions (366 days x 24 hours) to 366 dimensions (366 days). For simplicity and consistency, only the first PCA component was retained as the derived calendar data. Figure \ref{fig:preprocessing_meters} shows an example of implementing PCA on an electricity meter in an educational building, with a heatmap visualizing the intensity of energy use in daily values.

\begin{figure}
\begin{center}
\includegraphics[width=0.45\textwidth, trim= 0cm 0cm 0cm 0cm,clip]{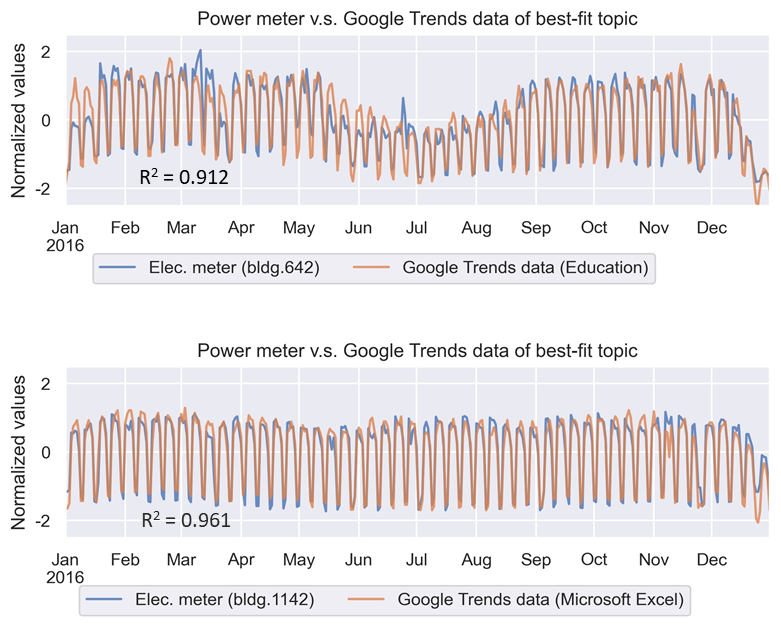}
\caption{Example of two building energy meters and their comparison to Google Trends topics}
\label{fig:meter_trends_example_comparisons}
\end{center}
\end{figure}

\subsubsection{Finding the best-fit topic for each power meter and building type}

Each power meter has its unique calendar pattern, so one of the focuses in this study is to find appropriate topics that would assist in modeling. The best-fit topic for each meter was found by calculating linear correlation to evaluate the similarity between the energy and Google Trends data. A past study also utilized linear correlation to select topics for improving the predictability of COVID-19 cases~\cite{Mavragani2020-yp}. Figure \ref{fig:meter_trends_example_comparisons} illustrates an example of two buildings. According to the calculation result of the correlation coefficient between the electricity meter and each topic, the search topic with the highest correlation to the calendar of the first building was \emph{Education} with an $R^2$ of 0.912. For the second building, the topic was \emph{Microsoft Excel} with an $R^2$ of 0.961.

\subsection{Prediction models to test the effectiveness of including Google Trends}
The baseline model in this study was chosen for its balanced performance between simplicity and accuracy, and it is a publicly-shared model in GEPIII on the Kaggle platform. This LightGBM-based model was developed using a 3-fold cross-validation method with basic data cleaning (e.g., abnormally extreme values and days-long constants were removed). The only difference between the proposed method and the baseline model is the additional feature of Google Trends data as the model settings and parameters were implemented in the same way. Table \ref{tab:model_parameter_overview} shows an overview of these two models.

In addition to evaluating the overall performance of the proposed method, the respective impacts of different day types are also considered to analyze further the benefits of Google Trend data in energy modeling. In this study, a majority of the power meters in the dataset are from university campuses. The academic calendar of universities was collected and manually labeled as (1) regular day, (2) public holiday, and (3) site-specific schedules. The definition of regular and public holidays day types were taken from online sources. Site-specific schedules are breaks and vacations of universities, which are unique to that site based on operational and academic calendars. Upon analyzing error reduction aggregated by different day types, the potential for Google Trends data to help automate the process of accounting for calendar data could be evaluated. This process might be beneficial, especially during the site-specific schedules, which are generally the most challenging period to predict.

\begin{table*}
\centering
\resizebox{0.9\textwidth}{!}{%
\begin{tabular}{lll}
\hline
\textbf{} & \textbf{Baseline model} & \textbf{Proposed method} \\ \hline
Name of notebook & KFold LightGBM -   without leak {[}1.062{]}* & - \\
Preprocessing & Remove outlier, Imputation & Remove outlier, Imputation \\
Features & \begin{tabular}[c]{@{}l@{}}Weather data, building meta, and  temporal \\ features (Hour of day and day of week)\end{tabular} & \begin{tabular}[c]{@{}l@{}}Weather data, building meta, temporal features \\ (Hour of day and day of week), and Google Trends data\end{tabular} \\
Number   of features & 11 & 11+1 \\ \hline
\multicolumn{3}{l}{*   Link of the notebook:   https://www.kaggle.com/teeyee314/kfold-lightgbm-without-leak-1-062}
\end{tabular}
}
\caption{Overview of the models developed to test the capability of the Google Trends topics to aid in the prediction of energy consumption}
\label{tab:model_parameter_overview}
\end{table*}

\subsection{Evaluation metrics}

The evaluation metric for this research is Root Mean Squared Logarithmic Error (RMSLE), which is consistent with ASHRAE's evaluation method in the Kaggle competition. If the data is skewed with extreme outliers, the error will be significantly amplified when using Root Mean Squared Error (RMSE) evaluation. Therefore, this metric was selected as it calculated the relative error between prediction and actual values while having robustness to the outliers. 

RMSLE is calculated according to Equation \ref{eqn:rmsle} with the following definitions:

\begin{equation}
\label{eqn:rmsle}
RMSLE = \sqrt{\frac{1}{n} \sum_{i=1}^n (\log(p_i + 1) - \log(a_i+1))^2 }
\end{equation}

\begin{itemize}
\item $n$ is the total number of observations
\item $pi$ is prediction of target
\item $a_i$ is the actual target for $i$
\item $log(x)$ is the natural logarithm of $x$
\end{itemize}

\section{Results}
\label{sec:results}

In this section, correlations between Google Trends data of topics and energy data and how they affect the prediction result are studied. Furthermore, to give a more intuitive understanding of RMSLE reduction, a benchmark of RMSLE collected from different levels of competitors in GEPIII is provided for comparison.

\subsection{Correlation results}

\begin{figure}
\begin{center}
\includegraphics[width=0.45\textwidth, trim= 0cm 0cm 0cm 0cm,clip]{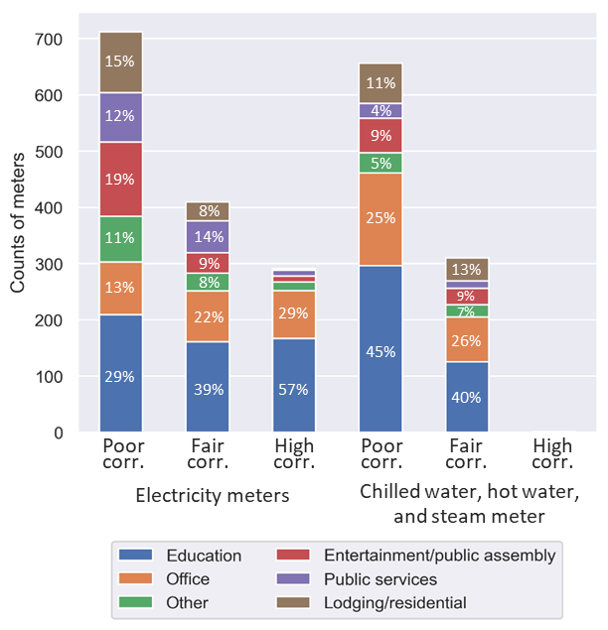}
\caption{Proportion of high, fair, and poor correlation amongst the building use types}
\label{fig:buildingtype_vs_correlation}
\end{center}
\end{figure}

Table \ref{tab:correlationAcrossMeterTypes} shows the results of the correlation analysis. Out of 2,380 meters, 293 were found to be highly correlated to the topics of Google Trends, and almost all of them are electricity-type meters. In contrast, most of the other three types of meters (chilled water, steam, and hot water) only have poor-correlated topics. The reason is that these three types are mainly related to environmental controls such as cooling and heating, so these meter readings are more dependent on weather variability, especially the outdoor temperature. Figure \ref{fig:buildingtype_vs_correlation} shows the distribution of correlation values according to building types and it can be seen that education and office building have the highest correlations. This result means that the energy use behaviors of these building types are more likely to find similar search trends with specific topics. For example, the two example buildings mentioned earlier are education and office buildings, which have high correlations with the topics \emph{Education}, and \emph{Microsoft Excel} respectively ($R^2>$0.8).

\begin{figure}
\begin{center}
\includegraphics[width=0.45\textwidth, trim= 0cm 0cm 0cm 0cm,clip]{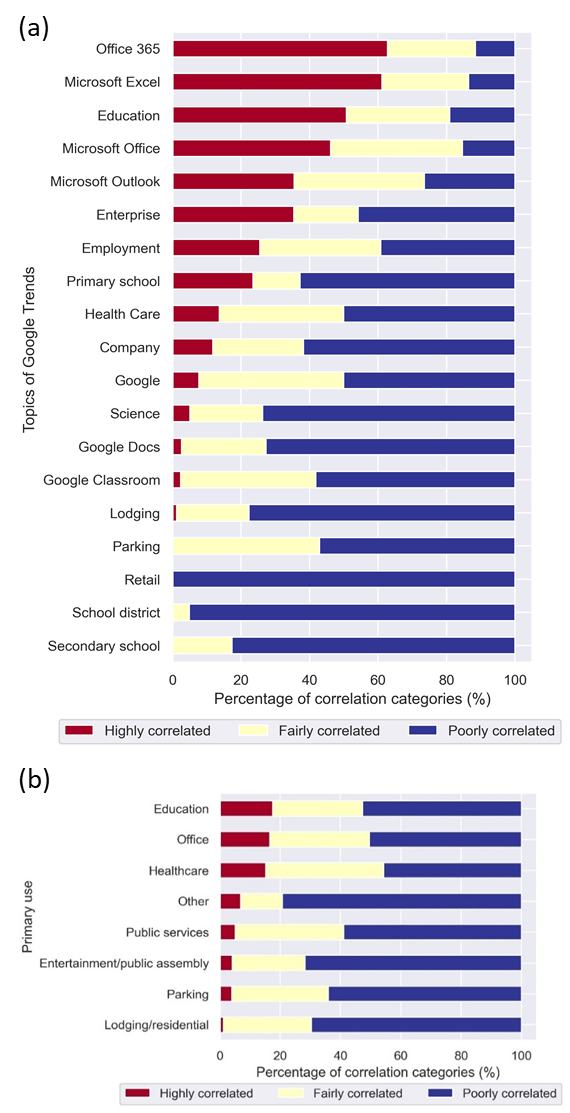}
\caption{Breakdown of the Google Trends topics according to their correlation with various building use types}
\label{fig:topics_vs_correlation}
\end{center}
\end{figure}

Figure \ref{fig:topics_vs_correlation}a shows that education and office-related topics have the highest percentage of high correlation to electricity meters. This result shows that Google Trends of these topics have more similarities to the time series of power meters, which may better explain building energy behavior. For example, office-related topics, such as \emph{Office 365}, \emph{Microsoft Excel}, \emph{Enterprise} and \emph{Employment}, have 30-60\% of high correlations with power meters. As for education-related topics, such as \emph{Education} and \emph{Primary school}, with roughly 20-50\% of highly correlated power meters. This situation is also reinforced in Figure \ref{fig:topics_vs_correlation}b, where \emph{Education} and \emph{Office} buildings happen to have the highest proportion of highly correlated meters among all building types.

\begin{table}
\centering
\resizebox{0.45\textwidth}{!}{%
\begin{tabular}{lccc}
\hline
 & \multicolumn{3}{c}{\textbf{Correlation with Google Trends}} \\
 & \textbf{Poor} ($R^2$\textless{}0.6) & \textbf{Fair} (0.6\textless{}$R^2$\textless{}0.8) & \textbf{High} ($R^2$\textgreater{}{}0.8) \\ \hline
Electricity & 712 (29.9\%) & 410 (17.2\%) & 292 (12.2\%) \\
Chilled water & 191 (8.0\%) & 307 (12.9\%) & 0 (0.0\%) \\
Steam & 323 (13.6\%) & 0 (0.0\%) & 0 (0.0\%) \\
Hot water & 142 (6.0\%) & 2 (0.1\%) & 1 (0.0\%) \\
Sum & 1368 (57.5\%) & 719 (30.2\%) & 293 (12.3\%) \\ \hline
\end{tabular}
}
\caption{Overview of correlations between power meters and Google Trends across meter types}
\label{tab:correlationAcrossMeterTypes}
\end{table}

\subsection{Effects on model prediction}

Table \ref{tab:error_overview_metertype} shows the comparison of the prediction error of the proposed method after adding Google Trends with the baseline model. It can be observed that a tendency for higher correlation of topics helps reduce more errors. In terms of electricity meters, using the method on the high, fair, and poor correlation categories of buildings could reduce RMSLE by 1.9\%, 1.2\%, and 1.0\%, respectively. If different day types are considered, the errors from public holidays and site-specific schedules can be reduced by 24\% and 3.5\% for highly correlated topics. In contrast, no significant error reduction is found on regular days. The other three types of meters (chilled water, hot water, and steam meter) are highly dependent on weather conditions due to their functionality for heating and cooling. Without fair and high correlated topics, the error of non-electricity meters even increased after adding low correlated topics, harming the accuracy of the model prediction.

Table \ref{tab:error_overview_topic} shows the prediction errors of meters that are further analyzed with highly correlated topics. It can be seen that most of the topics are related to office and education, and these highly correlated topics could contribute to the reduction of errors. For example, \emph{Education} could reduce the RMSLE of meters in the US by 5.7\% on average, especially for public holidays and site-specific schedules by 30\% and 8.3\%, respectively. Likewise, \emph{Primary school} also benefits the prediction during holidays and site-specific schedules, but the signal error in the test dataset causes the overall RMSLE to increase slightly. Office-related topics, including \emph{Enterprise}, \emph{Microsoft Excel} and \emph{Microsoft Word}, are also effective in reducing RMSLE by about 1.0 - 5.0\%. These cases show that Google Trends data with high correlation could serve as a good calendar feature for energy models. 

Table \ref{tab: error_benchmark} shows the comparison of the prediction results with the benchmark from the GEPIII competition. It is observed that higher correlated topics elevate the prediction performance to a higher level. After removing leak data in the test dataset, the average RMSLE was calculated for each medal level as a benchmark. Power meters with high-correlated topics have a 1.9\% error reduction in RMSLE, which is equivalent to the shift to the Top 5 prize-winning level of the GEPIII competition. As for fair and poor correlated topics, the error increased by +0.2\% and +0.9\%, respectively, causing no change to their medal level.

\begin{figure}
\begin{center}
\includegraphics[width=0.5\textwidth, trim= 0cm 0cm 0cm 0cm,clip]{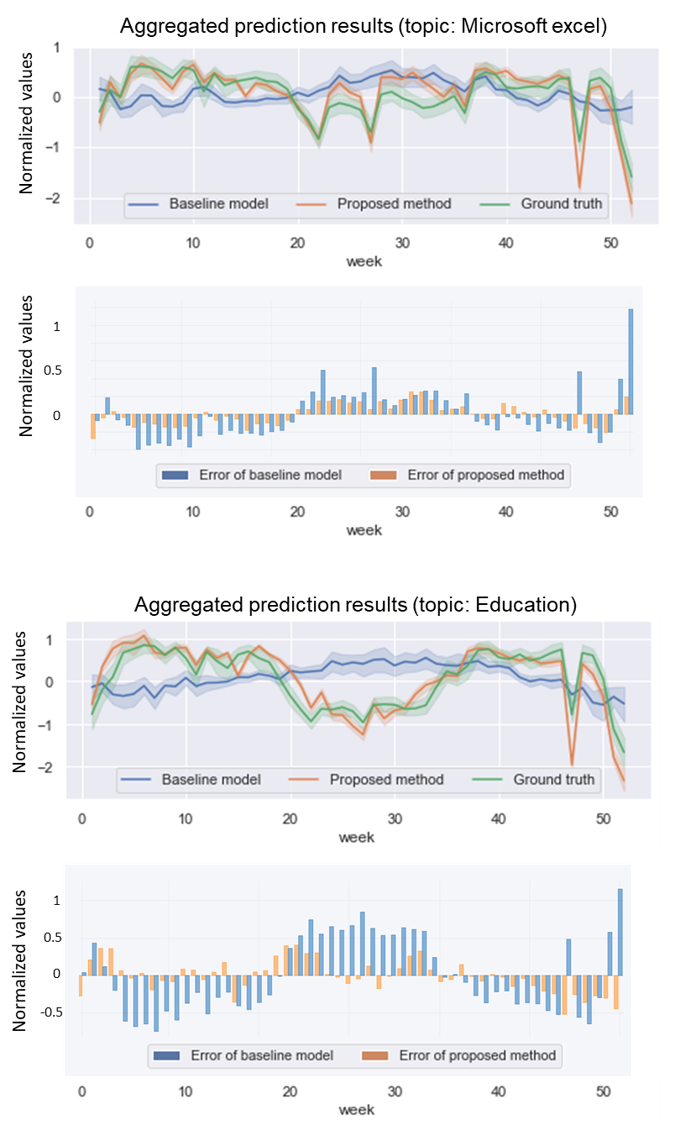}
\caption{Example of the error analysis of aggregated prediction results and the Google Trends topics that helped improve the model performance}
\label{fig:error_analysis_example2}
\end{center}
\end{figure}

\begin{table*}[]
\centering
\resizebox{0.9\textwidth}{!}{%
\begin{tabular}{llcccccccc}
\hline
\textbf{} & \textbf{} & \multicolumn{2}{c}{\textbf{RMSLE}} & \textbf{} & \multicolumn{4}{c}{\textbf{Change rate of error   (\%)}} & \textbf{} \\ \cline{3-4} \cline{6-9}
\textbf{Type} & \textbf{\begin{tabular}[c]{@{}l@{}}Correlation with \\ Google Trends\end{tabular}} & \textbf{\begin{tabular}[c]{@{}c@{}}Baseline \\ model\end{tabular}} & \textbf{\begin{tabular}[c]{@{}c@{}}Proposed \\ method\end{tabular}} & \textbf{} & \textbf{Regular} & \textbf{\begin{tabular}[c]{@{}c@{}}Public \\ holidays\end{tabular}} & \textbf{\begin{tabular}[c]{@{}c@{}}Site-specific \\ schedule\end{tabular}} & \textbf{Total} & \textbf{\begin{tabular}[c]{@{}c@{}}Count \\ of meters\end{tabular}} \\ \hline
Electricity & Poor & 0.475 & 0.470 &  & -0.5\% & -4.9\% & -1.5\% & -1.0\% & 712 \\
 & Fair & 0.510 & 0.505 &  & -0.2\% & -9.7\% & -1.8\% & -1.2\% & 410 \\
 & High & 0.549 & 0.539 &  & +0.3\% & -24.0\% & -3.5\% & -1.9\% & 291 \\
 &  &  &  &  &  &  &  &  &  \\
Chilled   Water & Poor & 1.226 & 1.243 &  & +1.5\% & +1.3\% & +1.4\% & +1.4\% & 191 \\
 & Fair & 0.949 & 0.950 &  & +0.4\% & +1.8\% & -0.6\% & +0.2\% & 307 \\
 &  &  &  &  &  &  &  &  &  \\
Steam & Poor & 1.146 & 1.150 &  & +0.1\% & +0.9\% & +1.1\% & +0.4\% & 323 \\
 & Fair & 1.055 & 0.941 &  & - & - & - & - & 1 \\
 &  &  &  &  &  &  &  &  &  \\
Hot water & Poor & 1.520 & 1.525 &  & +0.3\% & -0.6\% & +0.4\% & +0.3\% & 142 \\
 & Fair & 1.546 & 1.550 &  & - & - & - & - & 2 \\
 & High & 0.986 & 0.986 &  & - & - & - & - & 1 \\ \hline
\end{tabular}
}
\caption{Overview of the error analysis of the modeling experiment according to building meter type}
\label{tab:error_overview_metertype}
\end{table*}

\begin{table*}[]
\centering
\resizebox{0.9\textwidth}{!}{%
\begin{tabular}{llcclccccc}
\hline
\textbf{} & \textbf{} & \multicolumn{2}{c}{\textbf{RMSLE}} &  & \multicolumn{4}{c}{\textbf{Change rate of error   (\%)}} & \textbf{} \\ \cline{3-4} \cline{6-9}
\textbf{Country} & \textbf{\begin{tabular}[c]{@{}l@{}}Topic of \\ Google Trends\end{tabular}} & \textbf{\begin{tabular}[c]{@{}c@{}}Baseline \\ model\end{tabular}} & \textbf{\begin{tabular}[c]{@{}c@{}}Proposed \\ method\end{tabular}} &  & \textbf{Regular} & \textbf{\begin{tabular}[c]{@{}c@{}}Public \\ holidays\end{tabular}} & \textbf{\begin{tabular}[c]{@{}c@{}}Site-specific \\ schedule\end{tabular}} & \textbf{Total} & \textbf{\begin{tabular}[c]{@{}c@{}}Count \\ of meters\end{tabular}} \\ \hline
GB & Microsoft Excel & 1.523 & 1.522 &  & +0.6\% & -17.9\% & -0.4\% & -0.1\% & 10 \\
 & Office 365 & 1.754 & 1.743 &  & +1.0\% & -26.9\% & -2.8\% & -0.6\% & 26 \\
 & Primary school & 1.484 & 1.491 &  & +2.2\% & -23.9\% & -2.6\% & +0.5\% & 12 \\
 &  & \multicolumn{1}{l}{} & \multicolumn{1}{l}{} &  & \multicolumn{1}{l}{} & \multicolumn{1}{l}{} & \multicolumn{1}{l}{} & \multicolumn{1}{l}{} & \multicolumn{1}{l}{} \\
US & Education & 0.322 & 0.303 &  & -1.3\% & -30.0\% & -8.3\% & -5.7\% & 31 \\
 & Enterprise & 0.181 & 0.173 &  & +1.6\% & -30.0\% & -5.5\% & -4.2\% & 14 \\
 & Microsoft Excel & 0.277 & 0.266 &  & -1.3\% & -30.3\% & -3.8\% & -4.1\% & 51 \\
 & Microsoft Office & 0.425 & 0.419 &  & +0.5\% & -16.0\% & -3.2\% & -1.5\% & 12 \\
 & Microsoft Outlook & 0.307 & 0.296 &  & -2.0\% & -18.3\% & -2.7\% & -3.4\% & 55 \\
 & Office 365 & 0.239 & 0.236 &  & +0.6\% & -25.9\% & -0.7\% & -1.4\% & 23 \\ \hline
\end{tabular}
}
\caption{Overview of the change of error across topics}
\label{tab:error_overview_topic}
\end{table*}

\begin{table*}
\centering
\resizebox{0.9\textwidth}{!}{%
\begin{tabular}{llllllllll}
\hline
\textbf{} & \multicolumn{3}{c}{\textbf{RMSLE}} &  & \multicolumn{4}{c}{\textbf{\begin{tabular}[c]{@{}c@{}}Benchmark from GEPIII teams \\ (average RMSLE)\end{tabular}}} & \textbf{} \\ \cline{2-4} \cline{6-9}
\textbf{\begin{tabular}[c]{@{}l@{}}Correlation with \\ Google Trends\end{tabular}} & \textbf{\begin{tabular}[c]{@{}l@{}}Baseline \\ model\end{tabular}} & \textbf{\begin{tabular}[c]{@{}l@{}}Proposesd \\ method\end{tabular}} & \textbf{\begin{tabular}[c]{@{}l@{}}Change \\ rate\end{tabular}} &  & \textbf{\begin{tabular}[c]{@{}l@{}}Top 5 \\ winners\end{tabular}} &
\textbf{\begin{tabular}[c]{@{}l@{}}Gold \\ medal\end{tabular}} & \textbf{\begin{tabular}[c]{@{}l@{}}Silver \\ medal\end{tabular}} & \textbf{\begin{tabular}[c]{@{}l@{}}Bronze \\ medal\end{tabular}} &  \textbf{\begin{tabular}[c]{@{}l@{}}Count \\ of meters*\end{tabular}} \\ \hline

High ($R^2>$0.8) & 0.430 (Gold) & 0.422 (Top 5) & -1.9\% &  & 0.434 & 0.436 & 0.453 & 0.510 & 171 \\
Fair (0.8$>R^2>$0.6) & 1.002 (Bronze) & 1.004 (Bronze) & +0.2\% &  & 0.962 & 0.970 & 0.976 & 1.020 & 469 \\
Poor ($R^2<$0.6) & 1.182 (Bronze) & 1.193 (Bronze) & +0.9\% &  & 1.125 & 1.135 & 1.148 & 1.196 & 991 \\ \hline
\multicolumn{10}{l}{*Some of meters were removed because they were leak and became public to competitors in the GEPIII competition}
\end{tabular}
}

\caption{Error comparison between baseline and proposed method with benchmark from GEPIII}
\label{tab:error_benchmark}
\end{table*}

To further analyze the temporal change of error after adding Google Trends data, Figure \ref{fig:error_analysis_example2} shows aggregated prediction results and the temporal error in weekly values. Since office- and education-related topics have the most significant positive impact on the prediction accuracy, \emph{Microsoft Excel} and \emph{Education} in the US were selected as the topics in the example. For \emph{Microsoft Excel}, an office-related topic, the baseline model could not sufficiently predict the electricity consumption during holidays or vacations. In contrast, the proposed method with Google Trends can better predict energy use during holidays or vacations by using \emph{Microsoft Excel's} search volume as an additional feature. Topic \emph{Education} had similar positive benefits, with significant error reduction during site-specific schedules (e.g., winter and summer vacations). These visualization results show Google Trend's potential as an occupant-driven feature, helping the baseline model predict more accurately.

\begin{figure}
\begin{center}
\includegraphics[width=0.45\textwidth, trim= 0cm 0cm 0cm 0cm,clip]{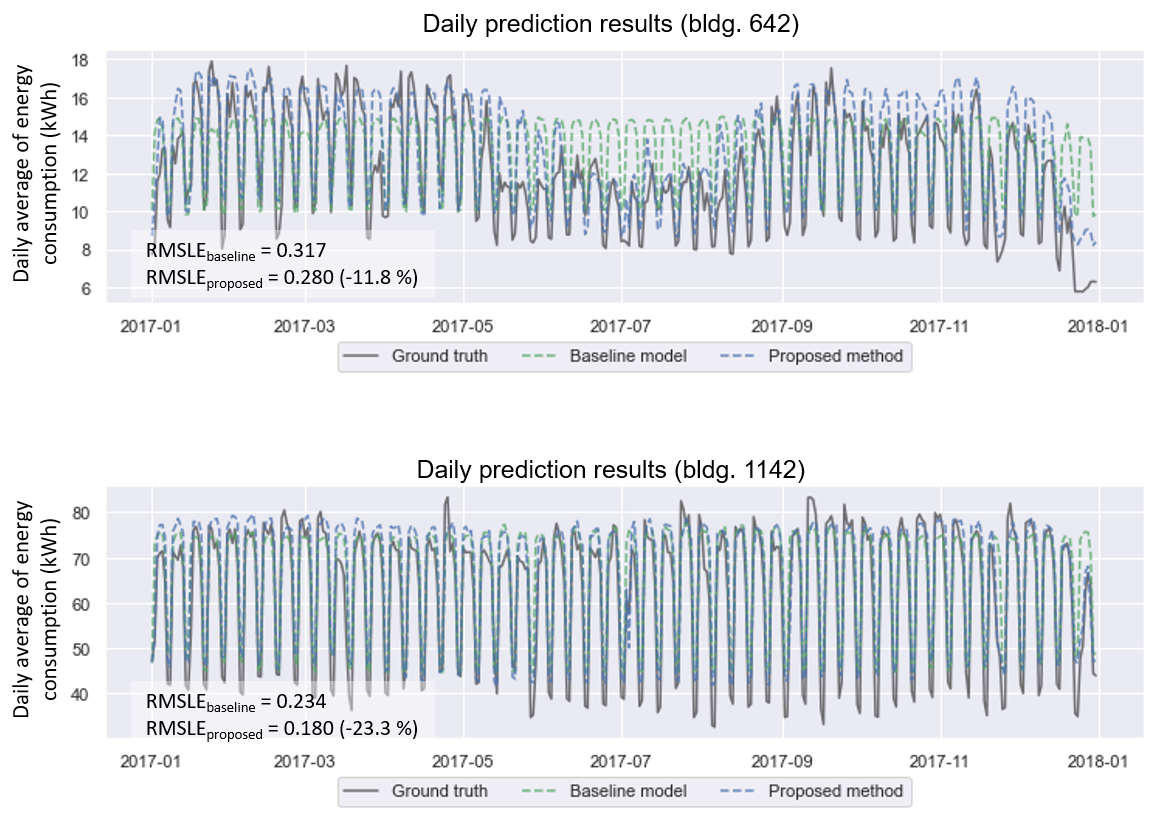}
\caption{Example of the error analysis of two meters and the Google Trends Topic that helped improve the model performance}
\label{fig:error_analysis_example1}
\end{center}
\end{figure}

Figure \ref{fig:error_analysis_example1} provides examples from two buildings, comparing the predictions before and after adding Google Trends. In the case of Building 642 (education-type building), the baseline model maintains a similar trend throughout the year, with significant errors during summer vacation and special holidays (e.g., Thanksgiving and Christmas). However, after adding Google Trends, the proposed method can predict energy values more accurately on these special days, and the RMSLE could be reduced from 0.317 to 0.280 (-11.8\%). Similarly, in offices such as Building 1142, the error of the baseline model mainly comes from national holidays, such as Thanksgiving and Christmas. With the help of Google Trends as a new feature, the proposed method can effectively reduce the RMSLE from 0.234 to 0.180 (-23.3\% ).

\section{Discussion}
\label{sec:discussion}

This analysis illustrates the ability for an innovative data source that has correlation and prediction power for energy consumption to be added to a process of modeling. 

\subsection{Reduction of error type frequency}
In previous work, the GEPIII competition was shown to have about 20\% of time with the classification of error~\cite{Miller2021-bf}. Errors were classified based on two main aspects: \emph{single-} or \emph{multi-building} (whether more than 33\% of buildings in a single site have simultaneous errors) and \emph{in-range} or \emph{out-of-range} (whether the scaled RMSLE is greater than 0.3). In this study, the proposed method of combining Google Trends into an energy model can specifically reduce the \emph{in-range} error of \emph{multiple buildings}. These non-extreme and cross-building errors mostly come from scheduling scenarios, such as breaks or holidays on university campuses. In most current energy models (including the GEPIII competition), the lack of schedules or calendars would cause simultaneous errors in a specific group of buildings. However, incorporating Google Trends data into energy modeling with correlation examination could reduce such errors through these alternative signals representing occupancy in buildings.

\subsection{General framework for testing influence of time-series data sources}
This framework opens doors for the screening and utilization of other data sets. Calendar data, in previous studies, have mainly used category variables or binary values to define day types as model inputs. This study provides a calendar feature with continuous values, the search volume of topics on the Google Trends platform, and model inputs that closely match energy usage behavior. Other data sources could be used in the same way to influence the prediction of building energy. WiFi connection data is beginning to emerge that can estimate the building's occupancy~\cite{Wang2017-tf, Zhan2021-jy, Nweye2020-yz}. Bluetooth Low Energy (BLE) systems can collect information about occupants' patterns of use of the building~\cite{Tekler2020-it, Jayathissa2020-pv}. Even text data sources from maintenance systems or emails could be extracted to understand the operations of the building~\cite{Miller2019-if, Gunay2019-un}.

\subsection{Potential impact on building analysis and control applications}
Improving building energy prediction for anomalous days can positively influence the implementation of various analytics and control approaches. For example, large-scale building energy benchmarking could utilize alternative data sources to increase the accuracy of determining buildings that are better or worse than their peers~\cite{Nutkiewicz2018-dk}. The method in this paper could influence the accuracy of short and medium-term prediction used for occupant-based building controls~\cite{Gunay2021-ly, Liu2021-pw}. Processing of data for calibration and the use in the simulation process could be improved~\cite{Chong2021-hs, Park2019-rn, Roth2020-bb}. The influence of predicting localized behavior better could better influence the classification of building use types~\cite{Quintana2021-dl}.

\subsection{Limitations of using Google Trends for energy prediction}
Out of 2380 power meters in the study, only 293 meters have high correlations with Google Trends data and significantly improve prediction performance. One of the main reasons is that around 50\% meters in the dataset, non-electricity meters, are more dependent on weather variables due to their role in heating and cooling the indoor environment. However, there are still more than 70\% electricity meters that don't have correlated trends from Google Trends topics. Besides, out of 39 building-related topics, less than ten were highly correlated with energy use ($R^2>$0.8). These highly correlated keywords can be mainly classified into office-related keywords (e.g. \emph{Enterprise}, \emph{Microsoft Excel}, etc.) and education-related keywords (e.g. \emph{Education}, \emph{Primary school}, etc.). This result indicates that Google Trends might only positively influence office and education building in the energy model. For other non-primary building types, like \emph{Parking} and \emph{Retail}, no highly correlated topics were found, resulting in no significant improvement for these cases. To fulfill these unsolved issues, more topics of Google Trends could be tested for finding more variation, or alternative datasets from other sources (e.g., WiFi and mobility data) could also be used for representing occupancy in buildings. There are numerous topics and terms that future researchers can explore to understand whether improvement is possible. It is possible even to try other online data sources such as Twitter or Linked-In that could contain signals that are proxies of behavior that influence building energy performance. 

Another limitation to consider is how Google Trends is affected by the size of the selected region, with a trade-off between region specificity and data quantity. Larger regions (e.g., countries) usually contain larger search volumes and have more stable and good data quality, but at the expense of losing region specificity. In contrast, smaller regions (e.g., cities) have unstable data quality due to less data, but the data could be much closer to the occupant behavior in the local area. This study uses country-wide Google Trends data as the calendar feature to ensure robustness with enough quantity. However, city- or state-specific events will not be specified in the country-wide Google Trends data, making it difficult to predict energy use for local events accurately.

\section{Conclusion}
\label{sec:conclusion}

This study shows a process of combining temporal data collected from Google Trends to predict energy consumption in buildings. It was found that the search volume of some building-related topics is highly correlated with building energy use from a subset of meters. The similarity between search volume and building energy use enables Google Trends data to help improve the prediction of energy models. In particular, during non-working periods (e.g., consecutive holidays and school breaks), Google Trends can effectively provide occupancy of different building types as a new feature. This insight effectively reduces the high cost needed in previous studies that required separately developing a model for each day type or manually collecting calendar data. 

This study also provides a simple but effective method for evaluating calendar features for energy prediction models. The best-fit topics can be found efficiently by calculating the linear correlation between Google Trends and energy data. In addition, according to the comparison of prediction results before and after adding Google Trends data, the data with high correlation ($R^2>$0.8) topics can significantly improve prediction results, comparable to top-5-winning solutions. On the other hand, however, Google Trends data with lower correlation risks increasing the prediction error. This framework could also be applied to similar data sets from other sources, such as transportation mobility data or WiFi connection data, and is expected to have similar effects in providing an occupant-driven feature, thereby improving energy prediction. The Google Trends topics used in this analysis include considerations for cross-language characteristics making the methodology applicable to energy data from building across sites or countries, giving the potential of incorporating occupant-driven data into energy modeling. 

\subsection{Reproducibility}
This analysis can be reproduced using the data and code from the following GitHub repository: \url{https://github.com/buds-lab/google-trends-for-buildings}. 

\section*{CRediT author statement}
\textbf{CF}: Conceptualization, Methodology, Software, Formal Analysis, Investigation, Data Curation, Visualization, Writing - Original Draft; \textbf{CM}: Conceptualization, Methodology, Resources, Writing - Reviewing \& Editing, Supervision, Project administration, Funding acquisition.

\section*{Funding}
The Singapore Ministry of Education (MOE) (R296000181133 and R296000214114) supported the development and implementation of this research.

\section*{Acknowledgements}
The authors would like to thank the team who developed and released the BDG2 data set and the technical aspects of the GEPIII competition, including (alphabetical order) Anjukan Kathirgamanathan, Bianca Bicchetti, Brodie Hobson, Forrest Meggers, June Young Park, Pandarasamy Arjunan, Paul Raftery, Zixiao Shi, and Zoltan Nagy. The authors would also like to thank the GEPIII planning committee members, including Anthony Fontanini, Chris Balbach, Jeff Haberl, and Krishnan Gowri. The ASHRAE organization is acknowledged for supporting the competition prize money and the Kaggle platform for hosting GEPIII as a non-profit competition.

\bibliographystyle{model1-num-names}
\bibliography{references}

\begin{thebibliography}{63}
\expandafter\ifx\csname natexlab\endcsname\relax\def\natexlab#1{#1}\fi
\providecommand{\bibinfo}[2]{#2}
\ifx\xfnm\relax \def\xfnm[#1]{\unskip,\space#1}\fi
\bibitem[{Wang et~al.(2020)Wang, Hong, and Piette}]{Wang2020-ky}
\bibinfo{author}{Z.~Wang}, \bibinfo{author}{T.~Hong}, \bibinfo{author}{M.~A.
  Piette},
\newblock \bibinfo{title}{Building thermal load prediction through shallow
  machine learning and deep learning},
\newblock \bibinfo{journal}{Applied Energy} \bibinfo{volume}{263}
  (\bibinfo{year}{2020}) \bibinfo{pages}{114683}.
\bibitem[{Fan et~al.(2021)Fan, Yan, Xiao, Li, An, and Kang}]{Fan2020-tu}
\bibinfo{author}{C.~Fan}, \bibinfo{author}{D.~Yan}, \bibinfo{author}{F.~Xiao},
  \bibinfo{author}{A.~Li}, \bibinfo{author}{J.~An}, \bibinfo{author}{X.~Kang},
\newblock \bibinfo{title}{Advanced data analytics for enhancing building
  performances: From data-driven to big data-driven approaches},
\newblock \bibinfo{journal}{Building Simulation} \bibinfo{volume}{14}
  (\bibinfo{year}{2021}) \bibinfo{pages}{3--24}.
\bibitem[{Li et~al.(2021)Li, Xiao, Fan, and Hu}]{Li2021-wb}
\bibinfo{author}{A.~Li}, \bibinfo{author}{F.~Xiao}, \bibinfo{author}{C.~Fan},
  \bibinfo{author}{M.~Hu},
\newblock \bibinfo{title}{Development of an {ANN-based} building energy model
  for information-poor buildings using transfer learning},
\newblock \bibinfo{journal}{Building Simulation} \bibinfo{volume}{14}
  (\bibinfo{year}{2021}) \bibinfo{pages}{89--101}.
\bibitem[{Granderson et~al.(2015)Granderson, Price, Jump, Addy, and
  Sohn}]{Granderson2015-ms}
\bibinfo{author}{J.~Granderson}, \bibinfo{author}{P.~N. Price},
  \bibinfo{author}{D.~Jump}, \bibinfo{author}{N.~Addy}, \bibinfo{author}{M.~D.
  Sohn},
\newblock \bibinfo{title}{Automated measurement and verification: Performance
  of public domain whole-building electric baseline models},
\newblock \bibinfo{journal}{Applied Energy} \bibinfo{volume}{144}
  (\bibinfo{year}{2015}) \bibinfo{pages}{106--113}.
\bibitem[{Granderson et~al.(2016)Granderson, Touzani, Custodio, Sohn, Jump, and
  Fernandes}]{Granderson2016-wq}
\bibinfo{author}{J.~Granderson}, \bibinfo{author}{S.~Touzani},
  \bibinfo{author}{C.~Custodio}, \bibinfo{author}{M.~D. Sohn},
  \bibinfo{author}{D.~Jump}, \bibinfo{author}{S.~Fernandes},
\newblock \bibinfo{title}{Accuracy of automated measurement and verification
  ({M\&V}) techniques for energy savings in commercial buildings},
\newblock \bibinfo{journal}{Applied Energy} \bibinfo{volume}{173}
  (\bibinfo{year}{2016}) \bibinfo{pages}{296--308}.
\bibitem[{Granderson et~al.(2017)Granderson, Touzani, Fernandes, and
  Taylor}]{Granderson2017-lm}
\bibinfo{author}{J.~Granderson}, \bibinfo{author}{S.~Touzani},
  \bibinfo{author}{S.~Fernandes}, \bibinfo{author}{C.~Taylor},
\newblock \bibinfo{title}{Application of automated measurement and verification
  to utility energy efficiency program data},
\newblock \bibinfo{journal}{Energy and Buildings} \bibinfo{volume}{142}
  (\bibinfo{year}{2017}) \bibinfo{pages}{191--199}.
\bibitem[{Chong et~al.(2017)Chong, Lam, Pozzi, and Yang}]{Chong2017-mk}
\bibinfo{author}{A.~Chong}, \bibinfo{author}{K.~P. Lam},
  \bibinfo{author}{M.~Pozzi}, \bibinfo{author}{J.~Yang},
\newblock \bibinfo{title}{Bayesian calibration of building energy models with
  large datasets},
\newblock \bibinfo{journal}{Energy and Buildings} \bibinfo{volume}{154}
  (\bibinfo{year}{2017}) \bibinfo{pages}{343--355}.
\bibitem[{Chong et~al.(2019)Chong, Xu, Chao, and Ngo}]{Chong2019-xh}
\bibinfo{author}{A.~Chong}, \bibinfo{author}{W.~Xu}, \bibinfo{author}{S.~Chao},
  \bibinfo{author}{N.-T. Ngo},
\newblock \bibinfo{title}{Continuous-time bayesian calibration of energy models
  using {BIM} and energy data},
\newblock \bibinfo{journal}{Energy and Buildings} \bibinfo{volume}{194}
  (\bibinfo{year}{2019}) \bibinfo{pages}{177--190}.
\bibitem[{Deb and Schlueter(2021)}]{Deb2021-py}
\bibinfo{author}{C.~Deb}, \bibinfo{author}{A.~Schlueter},
\newblock \bibinfo{title}{Review of data-driven energy modelling techniques for
  building retrofit},
\newblock \bibinfo{journal}{Renewable and Sustainable Energy Reviews}
  \bibinfo{volume}{144} (\bibinfo{year}{2021}) \bibinfo{pages}{110990}.
\bibitem[{Abdelrahman et~al.(2021)Abdelrahman, Zhan, Miller, and
  Chong}]{Abdelrahman2021-bk}
\bibinfo{author}{M.~M. Abdelrahman}, \bibinfo{author}{S.~Zhan},
  \bibinfo{author}{C.~Miller}, \bibinfo{author}{A.~Chong},
\newblock \bibinfo{title}{Data science for building energy efficiency: A
  comprehensive text-mining driven review of scientific literature},
\newblock \bibinfo{journal}{Energy and Buildings} \bibinfo{volume}{242}
  (\bibinfo{year}{2021}) \bibinfo{pages}{110885}.
\bibitem[{Kawashima et~al.(1995)Kawashima, Dorgan, and
  Mitchell}]{Kawashima1995-aw}
\bibinfo{author}{M.~Kawashima}, \bibinfo{author}{C.~E. Dorgan},
  \bibinfo{author}{J.~W. Mitchell}, \bibinfo{title}{Hourly thermal load
  prediction for the next 24 hours by {ARIMA}, {EWMA}, {LR} and an artificial
  neural network}, \bibinfo{type}{Technical Report}, ASHRAE,
  \bibinfo{year}{1995}.
\bibitem[{Ruch et~al.(1993)Ruch, Chen, Haberl, and Claridge}]{Ruch1993-bh}
\bibinfo{author}{D.~Ruch}, \bibinfo{author}{L.~Chen}, \bibinfo{author}{J.~S.
  Haberl}, \bibinfo{author}{D.~E. Claridge},
\newblock \bibinfo{title}{A {Change-Point} principal component analysis
  ({CP/PCA}) method for predicting energy usage in commercial buildings: The
  {PCA} model},
\newblock \bibinfo{journal}{Journal of Solar Energy Engineering}
  \bibinfo{volume}{115} (\bibinfo{year}{1993}) \bibinfo{pages}{77--84}.
\bibitem[{Gunay et~al.(2017)Gunay, Shen, and Newsham}]{Gunay2017-ke}
\bibinfo{author}{B.~Gunay}, \bibinfo{author}{W.~Shen},
  \bibinfo{author}{G.~Newsham},
\newblock \bibinfo{title}{Inverse blackbox modeling of the heating and cooling
  load in office buildings},
\newblock \bibinfo{journal}{Energy and Buildings} \bibinfo{volume}{142}
  (\bibinfo{year}{2017}) \bibinfo{pages}{200--210}.
\bibitem[{Neto and Fiorelli(2008)}]{Neto2008-bo}
\bibinfo{author}{A.~H. Neto}, \bibinfo{author}{F.~A.~S. Fiorelli},
\newblock \bibinfo{title}{Comparison between detailed model simulation and
  artificial neural network for forecasting building energy consumption},
\newblock \bibinfo{journal}{Energy and buildings} \bibinfo{volume}{40}
  (\bibinfo{year}{2008}) \bibinfo{pages}{2169--2176}.
\bibitem[{Moon et~al.(2019)Moon, Park, Rho, and Hwang}]{Moon2019-oy}
\bibinfo{author}{J.~Moon}, \bibinfo{author}{S.~Park}, \bibinfo{author}{S.~Rho},
  \bibinfo{author}{E.~Hwang},
\newblock \bibinfo{title}{A comparative analysis of artificial neural network
  architectures for building energy consumption forecasting},
\newblock \bibinfo{journal}{International Journal of Distributed Sensor
  Networks} \bibinfo{volume}{15} (\bibinfo{year}{2019})
  \bibinfo{pages}{1550147719877616}.
\bibitem[{Ahmad et~al.(2017)Ahmad, Mourshed, and Rezgui}]{Ahmad2017-jz}
\bibinfo{author}{M.~W. Ahmad}, \bibinfo{author}{M.~Mourshed},
  \bibinfo{author}{Y.~Rezgui},
\newblock \bibinfo{title}{Trees vs neurons: Comparison between random forest
  and {ANN} for high-resolution prediction of building energy consumption},
\newblock \bibinfo{journal}{Energy and Buildings} \bibinfo{volume}{147}
  (\bibinfo{year}{2017}) \bibinfo{pages}{77--89}.
\bibitem[{Brandi et~al.(2020)Brandi, Piscitelli, Martellacci, and
  Capozzoli}]{Brandi2020-ea}
\bibinfo{author}{S.~Brandi}, \bibinfo{author}{M.~S. Piscitelli},
  \bibinfo{author}{M.~Martellacci}, \bibinfo{author}{A.~Capozzoli},
\newblock \bibinfo{title}{Deep reinforcement learning to optimise indoor
  temperature control and heating energy consumption in buildings},
\newblock \bibinfo{journal}{Energy and Buildings} \bibinfo{volume}{224}
  (\bibinfo{year}{2020}) \bibinfo{pages}{110225}.
\bibitem[{Nichiforov et~al.(2018)Nichiforov, Stamatescu, Stamatescu, Calofir,
  Fagarasan, and Iliescu}]{Nichiforov2018-ne}
\bibinfo{author}{C.~Nichiforov}, \bibinfo{author}{G.~Stamatescu},
  \bibinfo{author}{I.~Stamatescu}, \bibinfo{author}{V.~Calofir},
  \bibinfo{author}{I.~Fagarasan}, \bibinfo{author}{S.~S. Iliescu},
\newblock \bibinfo{title}{Deep learning techniques for load forecasting in
  large commercial buildings},
\newblock in: \bibinfo{booktitle}{2018 22nd International Conference on System
  Theory, Control and Computing ({ICSTCC})}, \bibinfo{organization}{IEEE}, pp.
  \bibinfo{pages}{492--497}.
\bibitem[{Fan et~al.(2017)Fan, Xiao, and Zhao}]{Fan2017-ac}
\bibinfo{author}{C.~Fan}, \bibinfo{author}{F.~Xiao}, \bibinfo{author}{Y.~Zhao},
\newblock \bibinfo{title}{A short-term building cooling load prediction method
  using deep learning algorithms},
\newblock \bibinfo{journal}{Applied energy} \bibinfo{volume}{195}
  (\bibinfo{year}{2017}) \bibinfo{pages}{222--233}.
\bibitem[{Touzani et~al.(2018)Touzani, Granderson, and
  Fernandes}]{Touzani2018-lz}
\bibinfo{author}{S.~Touzani}, \bibinfo{author}{J.~Granderson},
  \bibinfo{author}{S.~Fernandes},
\newblock \bibinfo{title}{Gradient boosting machine for modeling the energy
  consumption of commercial buildings},
\newblock \bibinfo{journal}{Energy and Buildings} \bibinfo{volume}{158}
  (\bibinfo{year}{2018}) \bibinfo{pages}{1533--1543}.
\bibitem[{Amasyali and El-Gohary(2018)}]{Amasyali2018-wj}
\bibinfo{author}{K.~Amasyali}, \bibinfo{author}{N.~M. El-Gohary},
\newblock \bibinfo{title}{A review of data-driven building energy consumption
  prediction studies},
\newblock \bibinfo{journal}{Renewable and Sustainable Energy Reviews}
  \bibinfo{volume}{81} (\bibinfo{year}{2018}) \bibinfo{pages}{1192--1205}.
\bibitem[{Miller et~al.(2020)Miller, Arjunan, Kathirgamanathan, Fu, Roth, Park,
  Balbach, Gowri, Nagy, Fontanini et~al.}]{Miller2020-fo}
\bibinfo{author}{C.~Miller}, \bibinfo{author}{P.~Arjunan},
  \bibinfo{author}{A.~Kathirgamanathan}, \bibinfo{author}{C.~Fu},
  \bibinfo{author}{J.~Roth}, \bibinfo{author}{J.~Y. Park},
  \bibinfo{author}{C.~Balbach}, \bibinfo{author}{K.~Gowri},
  \bibinfo{author}{Z.~Nagy}, \bibinfo{author}{A.~D. Fontanini}, et~al.,
\newblock \bibinfo{title}{The ashrae great energy predictor iii competition:
  Overview and results},
\newblock \bibinfo{journal}{Science and Technology for the Built Environment}
  \bibinfo{volume}{26} (\bibinfo{year}{2020}) \bibinfo{pages}{1427--1447}.
\bibitem[{Haberl and Thamilseran(1998)}]{Haberl1998-du}
\bibinfo{author}{J.~S. Haberl}, \bibinfo{author}{S.~Thamilseran},
\newblock \bibinfo{title}{The great energy predictor shootout {II}},
\newblock \bibinfo{journal}{ASHRAE Journal} \bibinfo{volume}{40}
  (\bibinfo{year}{1998}) \bibinfo{pages}{49}.
\bibitem[{Kreider and Haberl(1994)}]{Kreider1994-dn}
\bibinfo{author}{J.~F. Kreider}, \bibinfo{author}{J.~S. Haberl},
  \bibinfo{title}{Predicting hourly building energy use: The great energy
  predictor shootout--Overview and discussion of results},
  \bibinfo{type}{Technical Report}, ASHRAE, \bibinfo{year}{1994}.
\bibitem[{Katipamula(1996)}]{Katipamula1996-et}
\bibinfo{author}{S.~Katipamula}, \bibinfo{title}{Great energy predictor
  shootout {II}: modeling energy use in large commercial buildings},
  \bibinfo{type}{Technical Report}, ASHRAE, \bibinfo{year}{1996}.
\bibitem[{Ohlsson et~al.(1994)Ohlsson, Peterson, Pi, Rognvaldsson, and
  Soderberg}]{Ohlsson1994-vl}
\bibinfo{author}{M.~B.~O. Ohlsson}, \bibinfo{author}{C.~O. Peterson},
  \bibinfo{author}{H.~Pi}, \bibinfo{author}{T.~S. Rognvaldsson},
  \bibinfo{author}{B.~P.~W. Soderberg},
\newblock \bibinfo{title}{Predicting system loads with artificial neural
  {Networks--Methods} and results from`` the great energy predictor
  shootout''},
\newblock \bibinfo{journal}{ASHRAE Transactions} \bibinfo{volume}{100}
  (\bibinfo{year}{1994}) \bibinfo{pages}{1063--1074}.
\bibitem[{Miller et~al.(2020)Miller, Kathirgamanathan, Picchetti, Arjunan,
  Park, Nagy, Raftery, Hobson, Shi, and Meggers}]{Miller2020-yc}
\bibinfo{author}{C.~Miller}, \bibinfo{author}{A.~Kathirgamanathan},
  \bibinfo{author}{B.~Picchetti}, \bibinfo{author}{P.~Arjunan},
  \bibinfo{author}{J.~Y. Park}, \bibinfo{author}{Z.~Nagy},
  \bibinfo{author}{P.~Raftery}, \bibinfo{author}{B.~W. Hobson},
  \bibinfo{author}{Z.~Shi}, \bibinfo{author}{F.~Meggers},
\newblock \bibinfo{title}{The building data genome project 2, energy meter data
  from the {ASHRAE} great energy predictor {III} competition},
\newblock \bibinfo{journal}{Scientific data} \bibinfo{volume}{7}
  (\bibinfo{year}{2020}) \bibinfo{pages}{1--13}.
\bibitem[{Miller(2019)}]{Miller2019-sg}
\bibinfo{author}{C.~Miller},
\newblock \bibinfo{title}{More buildings make more generalizable
  {Models---Benchmarking} prediction methods on open electrical meter data},
\newblock \bibinfo{journal}{Machine Learning and Knowledge Extraction}
  \bibinfo{volume}{1} (\bibinfo{year}{2019}) \bibinfo{pages}{974--993}.
\bibitem[{Miller et~al.(2021)Miller, Picchetti, Fu, and
  Pantelic}]{Miller2021-bf}
\bibinfo{author}{C.~Miller}, \bibinfo{author}{B.~Picchetti},
  \bibinfo{author}{C.~Fu}, \bibinfo{author}{J.~Pantelic},
  \bibinfo{title}{Limitations of machine learning for building energy
  prediction: {ASHRAE} great energy predictor {III} kaggle competition error
  analysis}, \bibinfo{year}{2021}.
\bibitem[{Touzani et~al.(2019)Touzani, Ravache, Crowe, and
  Granderson}]{Touzani2019-sz}
\bibinfo{author}{S.~Touzani}, \bibinfo{author}{B.~Ravache},
  \bibinfo{author}{E.~Crowe}, \bibinfo{author}{J.~Granderson},
\newblock \bibinfo{title}{Statistical change detection of building energy
  consumption: Applications to savings estimation},
\newblock \bibinfo{journal}{Energy and Buildings} \bibinfo{volume}{185}
  (\bibinfo{year}{2019}) \bibinfo{pages}{123--136}.
\bibitem[{Wi et~al.(2011)Wi, Joo, and Song}]{Wi2012-rv}
\bibinfo{author}{Y.~Wi}, \bibinfo{author}{S.~Joo}, \bibinfo{author}{K.~Song},
\newblock \bibinfo{title}{Holiday load forecasting using fuzzy polynomial
  regression with weather feature selection and adjustment},
\newblock \bibinfo{journal}{IEEE Transactions on Power Systems}
  \bibinfo{volume}{27} (\bibinfo{year}{2011}) \bibinfo{pages}{596--603}.
\bibitem[{Ebrahimi and Moshari(2013)}]{Ebrahimi2013-pk}
\bibinfo{author}{A.~Ebrahimi}, \bibinfo{author}{A.~Moshari},
\newblock \bibinfo{title}{Holidays short-term load forecasting using fuzzy
  improved similar day method},
\newblock \bibinfo{journal}{International transactions on electrical energy
  systems} \bibinfo{volume}{23} (\bibinfo{year}{2013})
  \bibinfo{pages}{1254--1271}.
\bibitem[{Ma et~al.(2017)Ma, Song, and Zhang}]{Ma2017-vk}
\bibinfo{author}{Z.~Ma}, \bibinfo{author}{J.~Song}, \bibinfo{author}{J.~Zhang},
\newblock \bibinfo{title}{Energy consumption prediction of air-conditioning
  systems in buildings by selecting similar days based on combined weights},
\newblock \bibinfo{journal}{Energy and Buildings} \bibinfo{volume}{151}
  (\bibinfo{year}{2017}) \bibinfo{pages}{157--166}.
\bibitem[{Dahl et~al.(2018)Dahl, Brun, Kirsebom, and Andresen}]{Dahl2018-ts}
\bibinfo{author}{M.~Dahl}, \bibinfo{author}{A.~Brun}, \bibinfo{author}{O.~S.
  Kirsebom}, \bibinfo{author}{G.~B. Andresen},
\newblock \bibinfo{title}{Improving {Short-Term} heat load forecasts with
  calendar and holiday data},
\newblock \bibinfo{journal}{Energies} \bibinfo{volume}{11}
  (\bibinfo{year}{2018}) \bibinfo{pages}{1678}.
\bibitem[{Zhang et~al.(2019)Zhang, Li, Zou, and Quiring}]{Zhang2019-dr}
\bibinfo{author}{N.~Zhang}, \bibinfo{author}{Z.~Li}, \bibinfo{author}{X.~Zou},
  \bibinfo{author}{S.~M. Quiring},
\newblock \bibinfo{title}{Comparison of three short-term load forecast models
  in southern california},
\newblock \bibinfo{journal}{Energy} \bibinfo{volume}{189}
  (\bibinfo{year}{2019}) \bibinfo{pages}{116358}.
\bibitem[{Zhang et~al.(2018)Zhang, Bambrick, Mengersen, Tong, and
  Hu}]{Zhang2018-av}
\bibinfo{author}{Y.~Zhang}, \bibinfo{author}{H.~Bambrick},
  \bibinfo{author}{K.~Mengersen}, \bibinfo{author}{S.~Tong},
  \bibinfo{author}{W.~Hu},
\newblock \bibinfo{title}{Using google trends and ambient temperature to
  predict seasonal influenza outbreaks},
\newblock \bibinfo{journal}{Environment international} \bibinfo{volume}{117}
  (\bibinfo{year}{2018}) \bibinfo{pages}{284--291}.
\bibitem[{Nuti et~al.(2014)Nuti, Wayda, Ranasinghe, Wang, Dreyer, Chen, and
  Murugiah}]{Nuti2014-bu}
\bibinfo{author}{S.~V. Nuti}, \bibinfo{author}{B.~Wayda},
  \bibinfo{author}{I.~Ranasinghe}, \bibinfo{author}{S.~Wang},
  \bibinfo{author}{R.~P. Dreyer}, \bibinfo{author}{S.~I. Chen},
  \bibinfo{author}{K.~Murugiah},
\newblock \bibinfo{title}{The use of google trends in health care research: a
  systematic review},
\newblock \bibinfo{journal}{PloS one} \bibinfo{volume}{9}
  (\bibinfo{year}{2014}) \bibinfo{pages}{e109583}.
\bibitem[{Mavragani and Gkillas(2020)}]{Mavragani2020-yp}
\bibinfo{author}{A.~Mavragani}, \bibinfo{author}{K.~Gkillas},
\newblock \bibinfo{title}{{COVID-19} predictability in the united states using
  google trends time series},
\newblock \bibinfo{journal}{Scientific reports} \bibinfo{volume}{10}
  (\bibinfo{year}{2020}) \bibinfo{pages}{1--12}.
\bibitem[{Preis et~al.(2013)Preis, Moat, and Stanley}]{Preis2013-du}
\bibinfo{author}{T.~Preis}, \bibinfo{author}{H.~S. Moat},
  \bibinfo{author}{H.~E. Stanley},
\newblock \bibinfo{title}{Quantifying trading behavior in financial markets
  using google trends},
\newblock \bibinfo{journal}{Scientific reports} \bibinfo{volume}{3}
  (\bibinfo{year}{2013}) \bibinfo{pages}{1--6}.
\bibitem[{Woo and Owen(2019)}]{Woo2018-ej}
\bibinfo{author}{J.~Woo}, \bibinfo{author}{A.~L. Owen},
\newblock \bibinfo{title}{Forecasting private consumption with google trends
  data},
\newblock \bibinfo{journal}{Journal of Forecasting} \bibinfo{volume}{38}
  (\bibinfo{year}{2019}) \bibinfo{pages}{81--91}.
\bibitem[{Carri{\`e}re-Swallow and Labb{\'e}(2013)}]{Carriere-Swallow2013-fu}
\bibinfo{author}{Y.~Carri{\`e}re-Swallow}, \bibinfo{author}{F.~Labb{\'e}},
\newblock \bibinfo{title}{Nowcasting with google trends in an emerging market},
\newblock \bibinfo{journal}{Journal of Forecasting} \bibinfo{volume}{32}
  (\bibinfo{year}{2013}) \bibinfo{pages}{289--298}.
\bibitem[{Vosen and Schmidt(2011)}]{Vosen2011-oo}
\bibinfo{author}{S.~Vosen}, \bibinfo{author}{T.~Schmidt},
\newblock \bibinfo{title}{Forecasting private consumption: survey-based
  indicators vs. google trends},
\newblock \bibinfo{journal}{Journal of forecasting} \bibinfo{volume}{30}
  (\bibinfo{year}{2011}) \bibinfo{pages}{565--578}.
\bibitem[{Clark et~al.(2019)Clark, Wilkins, Dagan, Powell, Sharp, and
  Hillis}]{Clark2019-bg}
\bibinfo{author}{M.~Clark}, \bibinfo{author}{E.~J. Wilkins},
  \bibinfo{author}{D.~T. Dagan}, \bibinfo{author}{R.~Powell},
  \bibinfo{author}{R.~L. Sharp}, \bibinfo{author}{V.~Hillis},
\newblock \bibinfo{title}{Bringing forecasting into the future: Using google to
  predict visitation in {U.S}. national parks},
\newblock \bibinfo{journal}{Journal of environmental management}
  \bibinfo{volume}{243} (\bibinfo{year}{2019}) \bibinfo{pages}{88--94}.
\bibitem[{{\"O}nder(2017)}]{Onder2017-jo}
\bibinfo{author}{I.~{\"O}nder},
\newblock \bibinfo{title}{Forecasting tourism demand with google trends:
  Accuracy comparison of countries versus cities},
\newblock \bibinfo{journal}{International Journal of Tourism Research}
  \bibinfo{volume}{19} (\bibinfo{year}{2017}) \bibinfo{pages}{648--660}.
\bibitem[{Dergiades et~al.(2018)Dergiades, Mavragani, and
  Pan}]{Dergiades2018-ml}
\bibinfo{author}{T.~Dergiades}, \bibinfo{author}{E.~Mavragani},
  \bibinfo{author}{B.~Pan},
\newblock \bibinfo{title}{Google trends and tourists' arrivals: Emerging biases
  and proposed corrections},
\newblock \bibinfo{journal}{Tourism Management} \bibinfo{volume}{66}
  (\bibinfo{year}{2018}) \bibinfo{pages}{108--120}.
\bibitem[{Happle et~al.(2020)Happle, Fonseca, and Schlueter}]{Happle2020-el}
\bibinfo{author}{G.~Happle}, \bibinfo{author}{J.~A. Fonseca},
  \bibinfo{author}{A.~Schlueter},
\newblock \bibinfo{title}{Context-specific urban occupancy modeling using
  location-based services data},
\newblock \bibinfo{journal}{Building and Environment} \bibinfo{volume}{175}
  (\bibinfo{year}{2020}) \bibinfo{pages}{106803}.
\bibitem[{Barbour et~al.(2019)Barbour, Davila, Gupta, Reinhart, Kaur, and
  Gonz{\'a}lez}]{Barbour2019-sc}
\bibinfo{author}{E.~Barbour}, \bibinfo{author}{C.~C. Davila},
  \bibinfo{author}{S.~Gupta}, \bibinfo{author}{C.~Reinhart},
  \bibinfo{author}{J.~Kaur}, \bibinfo{author}{M.~C. Gonz{\'a}lez},
\newblock \bibinfo{title}{Planning for sustainable cities by estimating
  building occupancy with mobile phones},
\newblock \bibinfo{journal}{Nature communications} \bibinfo{volume}{10}
  (\bibinfo{year}{2019}) \bibinfo{pages}{1--10}.
\bibitem[{Kang et~al.(2019)Kang, Yan, Sun, Jin, and Xu}]{Kang_undated-hd}
\bibinfo{author}{X.~Kang}, \bibinfo{author}{D.~Yan}, \bibinfo{author}{H.~Sun},
  \bibinfo{author}{Y.~Jin}, \bibinfo{author}{P.~Xu},
\newblock \bibinfo{title}{An approach for obtaining and extracting occupancy
  patterns in buildings based on mobile positioning data},
\newblock in: \bibinfo{booktitle}{Proceeding of IBPSA 2019 Conference}.
\bibitem[{Wold et~al.(1987)Wold, Esbensen, and Geladi}]{wold1987principal}
\bibinfo{author}{S.~Wold}, \bibinfo{author}{K.~Esbensen},
  \bibinfo{author}{P.~Geladi},
\newblock \bibinfo{title}{Principal component analysis},
\newblock \bibinfo{journal}{Chemometrics and intelligent laboratory systems}
  \bibinfo{volume}{2} (\bibinfo{year}{1987}) \bibinfo{pages}{37--52}.
\bibitem[{Wang and Shao(2017)}]{Wang2017-tf}
\bibinfo{author}{Y.~Wang}, \bibinfo{author}{L.~Shao},
\newblock \bibinfo{title}{Understanding occupancy pattern and improving
  building energy efficiency through {Wi-Fi} based indoor positioning},
\newblock \bibinfo{journal}{Building and Environment} \bibinfo{volume}{114}
  (\bibinfo{year}{2017}) \bibinfo{pages}{106--117}.
\bibitem[{Zhan and Chong(2021)}]{Zhan2021-jy}
\bibinfo{author}{S.~Zhan}, \bibinfo{author}{A.~Chong},
\newblock \bibinfo{title}{Building occupancy and energy consumption: Case
  studies across building types},
\newblock \bibinfo{journal}{Energy and Built Environment} \bibinfo{volume}{2}
  (\bibinfo{year}{2021}) \bibinfo{pages}{167--174}.
\bibitem[{Nweye and Nagy(2020)}]{Nweye2020-yz}
\bibinfo{author}{K.~Nweye}, \bibinfo{author}{Z.~Nagy},
\newblock \bibinfo{title}{{HVAC} scheduling based on {Wi-Fi} derived
  occupancy},
\newblock in: \bibinfo{booktitle}{Proceedings of the 7th {ACM} International
  Conference on Systems for {Energy-Efficient} Buildings, Cities, and
  Transportation}, BuildSys '20, \bibinfo{publisher}{Association for Computing
  Machinery}, \bibinfo{address}{New York, NY, USA}, \bibinfo{year}{2020}, pp.
  \bibinfo{pages}{340--341}.
\bibitem[{Tekler et~al.(2020)Tekler, Low, Gunay, Andersen, and
  Blessing}]{Tekler2020-it}
\bibinfo{author}{Z.~D. Tekler}, \bibinfo{author}{R.~Low},
  \bibinfo{author}{B.~Gunay}, \bibinfo{author}{R.~K. Andersen},
  \bibinfo{author}{L.~Blessing},
\newblock \bibinfo{title}{A scalable bluetooth low energy approach to identify
  occupancy patterns and profiles in office spaces},
\newblock \bibinfo{journal}{Building and Environment} \bibinfo{volume}{171}
  (\bibinfo{year}{2020}) \bibinfo{pages}{106681}.
\bibitem[{Jayathissa et~al.(2020)Jayathissa, Quintana, Abdelrahman, and
  Miller}]{Jayathissa2020-pv}
\bibinfo{author}{P.~Jayathissa}, \bibinfo{author}{M.~Quintana},
  \bibinfo{author}{M.~Abdelrahman}, \bibinfo{author}{C.~Miller},
\newblock \bibinfo{title}{{Humans-as-a-Sensor} for {Buildings---Intensive}
  longitudinal indoor comfort models},
\newblock \bibinfo{journal}{Buildings} \bibinfo{volume}{10}
  (\bibinfo{year}{2020}) \bibinfo{pages}{174}.
\bibitem[{Miller et~al.(2019)Miller, Quintana, and Glazer}]{Miller2019-if}
\bibinfo{author}{C.~Miller}, \bibinfo{author}{M.~Quintana},
  \bibinfo{author}{J.~Glazer},
\newblock \bibinfo{title}{Twenty years of building performance analysis trends:
  A topic modeling analysis of the {Bldg-Sim} email list archive},
\newblock in: \bibinfo{booktitle}{Building Simulation Conference Proceedings},
  pp. \bibinfo{pages}{Pages 1522--1529}.
\bibitem[{Gunay et~al.(2019)Gunay, Shen, and Yang}]{Gunay2019-un}
\bibinfo{author}{H.~B. Gunay}, \bibinfo{author}{W.~Shen},
  \bibinfo{author}{C.~Yang},
\newblock \bibinfo{title}{Text-mining building maintenance work orders for
  component fault frequency},
\newblock \bibinfo{journal}{Building Research \& Information}
  \bibinfo{volume}{47} (\bibinfo{year}{2019}) \bibinfo{pages}{518--533}.
\bibitem[{Nutkiewicz et~al.(2018)Nutkiewicz, Yang, and
  Jain}]{Nutkiewicz2018-dk}
\bibinfo{author}{A.~Nutkiewicz}, \bibinfo{author}{Z.~Yang},
  \bibinfo{author}{R.~K. Jain},
\newblock \bibinfo{title}{Data-driven urban energy simulation ({DUE-S)}: A
  framework for integrating engineering simulation and machine learning methods
  in a multi-scale urban energy modeling workflow},
\newblock \bibinfo{journal}{Applied energy} \bibinfo{volume}{225}
  (\bibinfo{year}{2018}) \bibinfo{pages}{1176--1189}.
\bibitem[{Gunay et~al.(2021)Gunay, Nagy, Miller, Ouf, and Dong}]{Gunay2021-ly}
\bibinfo{author}{B.~Gunay}, \bibinfo{author}{Z.~Nagy},
  \bibinfo{author}{C.~Miller}, \bibinfo{author}{M.~Ouf},
  \bibinfo{author}{B.~Dong},
\newblock \bibinfo{title}{Using {Occupant-Centric} control for commercial
  {HVAC} systems},
\newblock \bibinfo{journal}{ASHRAE Journal} \bibinfo{volume}{63}
  (\bibinfo{year}{2021}) \bibinfo{pages}{30--40}.
\bibitem[{Liu et~al.(2021)Liu, Lee, Bilionis, Karava, Joe, and
  Sadeghi}]{Liu2021-pw}
\bibinfo{author}{X.~Liu}, \bibinfo{author}{S.~Lee},
  \bibinfo{author}{I.~Bilionis}, \bibinfo{author}{P.~Karava},
  \bibinfo{author}{J.~Joe}, \bibinfo{author}{S.~A. Sadeghi},
\newblock \bibinfo{title}{A user-interactive system for smart thermal
  environment control in office buildings},
\newblock \bibinfo{journal}{Applied Energy} \bibinfo{volume}{298}
  (\bibinfo{year}{2021}) \bibinfo{pages}{117005}.
\bibitem[{Chong et~al.(2021)Chong, Augenbroe, and Yan}]{Chong2021-hs}
\bibinfo{author}{A.~Chong}, \bibinfo{author}{G.~Augenbroe},
  \bibinfo{author}{D.~Yan},
\newblock \bibinfo{title}{Occupancy data at different spatial resolutions:
  Building energy performance and model calibration},
\newblock \bibinfo{journal}{Applied Energy} \bibinfo{volume}{286}
  (\bibinfo{year}{2021}) \bibinfo{pages}{116492}.
\bibitem[{Park et~al.(2019)Park, Miller, and Nagy}]{Park2019-rn}
\bibinfo{author}{J.~Y. Park}, \bibinfo{author}{C.~Miller},
  \bibinfo{author}{Z.~Nagy},
\newblock \bibinfo{title}{A {Data-Driven} load shape profile based building
  benchmarking: Comparing doe reference buildings with a large metering
  dataset},
\newblock in: \bibinfo{booktitle}{Building Simulation Conference Proceedings},
  pp. \bibinfo{pages}{Pages 4282--4289}.
\bibitem[{Roth et~al.(2020)Roth, Martin, Miller, and Jain}]{Roth2020-bb}
\bibinfo{author}{J.~Roth}, \bibinfo{author}{A.~Martin},
  \bibinfo{author}{C.~Miller}, \bibinfo{author}{R.~K. Jain},
\newblock \bibinfo{title}{{SynCity}: Using open data to create a synthetic city
  of hourly building energy estimates by integrating data-driven and
  physics-based methods},
\newblock \bibinfo{journal}{Applied Energy} \bibinfo{volume}{280}
  (\bibinfo{year}{2020}) \bibinfo{pages}{115981}.
\bibitem[{Quintana et~al.(2021)Quintana, Arjunan, and Miller}]{Quintana2021-dl}
\bibinfo{author}{M.~Quintana}, \bibinfo{author}{P.~Arjunan},
  \bibinfo{author}{C.~Miller},
\newblock \bibinfo{title}{Islands of misfit buildings: Detecting
  uncharacteristic electricity use behavior using load shape clustering},
\newblock \bibinfo{journal}{Building Simulation} \bibinfo{volume}{14}
  (\bibinfo{year}{2021}) \bibinfo{pages}{119--130}.

\end{thebibliography}

\end{document}